\documentclass[10pt,twocolumn,letterpaper]{article}

\usepackage{graphicx}  
\usepackage{bm} 
\usepackage{multirow}
\usepackage{bbding}
\usepackage{colortbl}
\usepackage{amsmath}
\usepackage{graphicx}
\usepackage{changepage}
\usepackage{booktabs}
\usepackage[table]{xcolor}
\usepackage{pifont}
\usepackage{algorithm}
\usepackage{algorithmic}

\newcommand{\INPUT}{\REQUIRE}
\newcommand{\OUTPUT}{\ENSURE}

\usepackage{cvpr}              
\definecolor{cvprblue}{rgb}{0.21,0.49,0.74}
\usepackage[pagebackref,breaklinks,colorlinks,allcolors=cvprblue]{hyperref}

\def\paperID{} 
\def\confName{CVPR}
\def\confYear{2026}

\title{Missing No More: Dictionary-Guided Cross-Modal Image Fusion under Missing Infrared}

\author{%
	Yafei Zhang\textsuperscript{1}, %
	Meng Ma\textsuperscript{1}, %
	Huafeng Li\textsuperscript{1}\thanks{Corresponding author.}, %
	Yu Liu\textsuperscript{2}, %
	\\
	\textsuperscript{1}Faculty of Information Engineering and Automation, Kunming University of Science and Technology,\\
	\textsuperscript{2}Department of Biomedical Engineering, Hefei University of Technology\\
}

\begin{document}
\maketitle

\begin{abstract}
	Infrared-visible (IR-VIS) image fusion is vital for perception and security, yet most methods rely on the availability of both modalities during training and inference. When the infrared modality is absent, pixel-space generative substitutes become hard to control and inherently lack interpretability. We address missing-IR fusion by proposing a dictionary-guided, coefficient-domain framework built upon a shared convolutional dictionary. The pipeline comprises three key components: (1) Joint Shared-dictionary Representation Learning (JSRL) learns a unified and interpretable atom space shared by both IR and VIS modalities; (2) VIS-Guided IR Inference (VGII) transfers VIS coefficients to pseudo-IR coefficients in the coefficient domain and performs a one-step closed-loop refinement guided by a frozen large language model as a weak semantic prior; and (3) Adaptive Fusion via Representation Inference (AFRI) merges VIS structures and inferred IR cues at the atom level through window attention and convolutional mixing, followed by reconstruction with the shared dictionary. This \emph{encode$\rightarrow$transfer$\rightarrow$fuse$\rightarrow$reconstruct} pipeline avoids uncontrolled pixel-space generation while ensuring prior preservation within interpretable dictionary-coefficient representation. Experiments under missing-IR settings demonstrate consistent improvements in perceptual quality and downstream detection performance. To our knowledge, this represents the first framework that jointly learns a shared dictionary and performs coefficient-domain inference–fusion to tackle missing-IR fusion. The source code is publicly available at \href{https://github.com/harukiv/DCMIF}{\textcolor{blue}{https://github.com/harukiv/DCMIF}}.
	
\end{abstract}\vspace{-5mm}

\section{Introduction}
\label{sec:intro}
Infrared-visible (IR-VIS) image fusion is critical for robust perception in surveillance, robotics, and autonomous systems. Most existing approaches assume both modalities are available during training and inference, and adopt end-to-end designs—CNNs \cite{4,6,7}, CNN-Transformer hybrids \cite{3,48,31}, GANs \cite{8,42} , or diffusion models \cite{9,46}—to directly regress a fused image from paired inputs. While effective under standard conditions, this assumption often breaks in realistic missing-modality scenarios (e.g., VIS-only at test time). Moreover, pixel-space black-box generation offers weak physical consistency and interpretability, often resulting in unstable thermal cue completion, structural detail loss, or hallucinated patterns.

\begin{figure}[t!]
	\centering
	\includegraphics[width=0.425\textwidth]{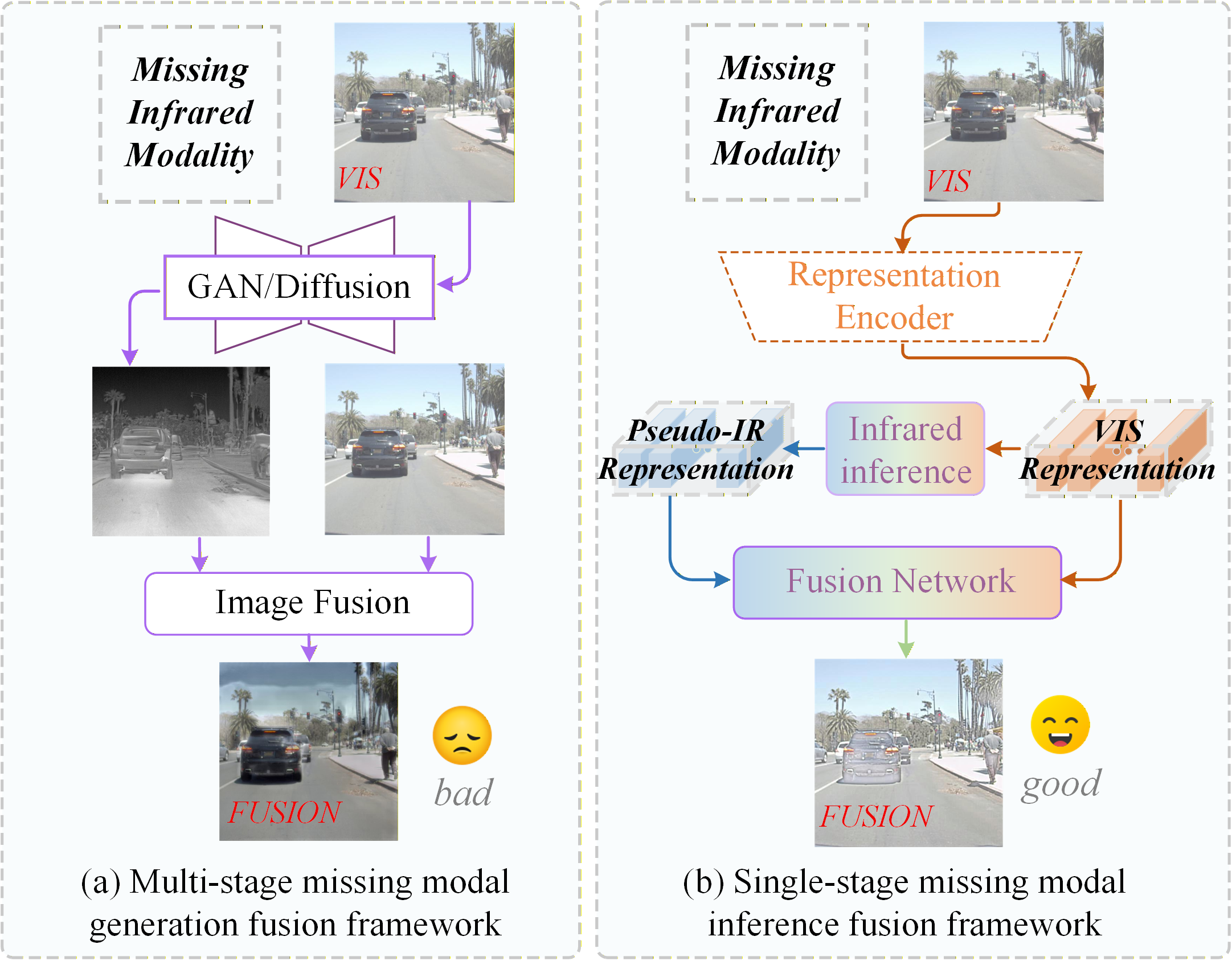}\vspace{-2mm}
	\caption{Comparison between existing methods and ours. For infrared-missing cross-modal fusion, (a) existing methods adopt a multi-stage framework that generates infrared images before fusion, while (b) we propose a single-stage framework that directly infers infrared features from visible images without generating infrared images.}\vspace{-5mm}
	\label{fig1}
\end{figure}

We address IR-VIS fusion under missing-IR conditions by predicting infrared cues from visible information and fusing them with the visible image, thereby enhancing perceptual quality and improving performance in downstream tasks such as object detection and semantic segmentation. As shown in Figure \ref{fig1}, rather than directly generating IR in pixel space, we map both modalities into a unified dictionary-coefficient space to perform interpretable inference and fusion in the coefficient domain. This anchors data consistency and prior constraints at the atom-coefficient level, effectively avoiding the brittleness of pixel-space generation.

\begin{figure*}[t!]
	\centering
	\includegraphics[width=0.9\textwidth]{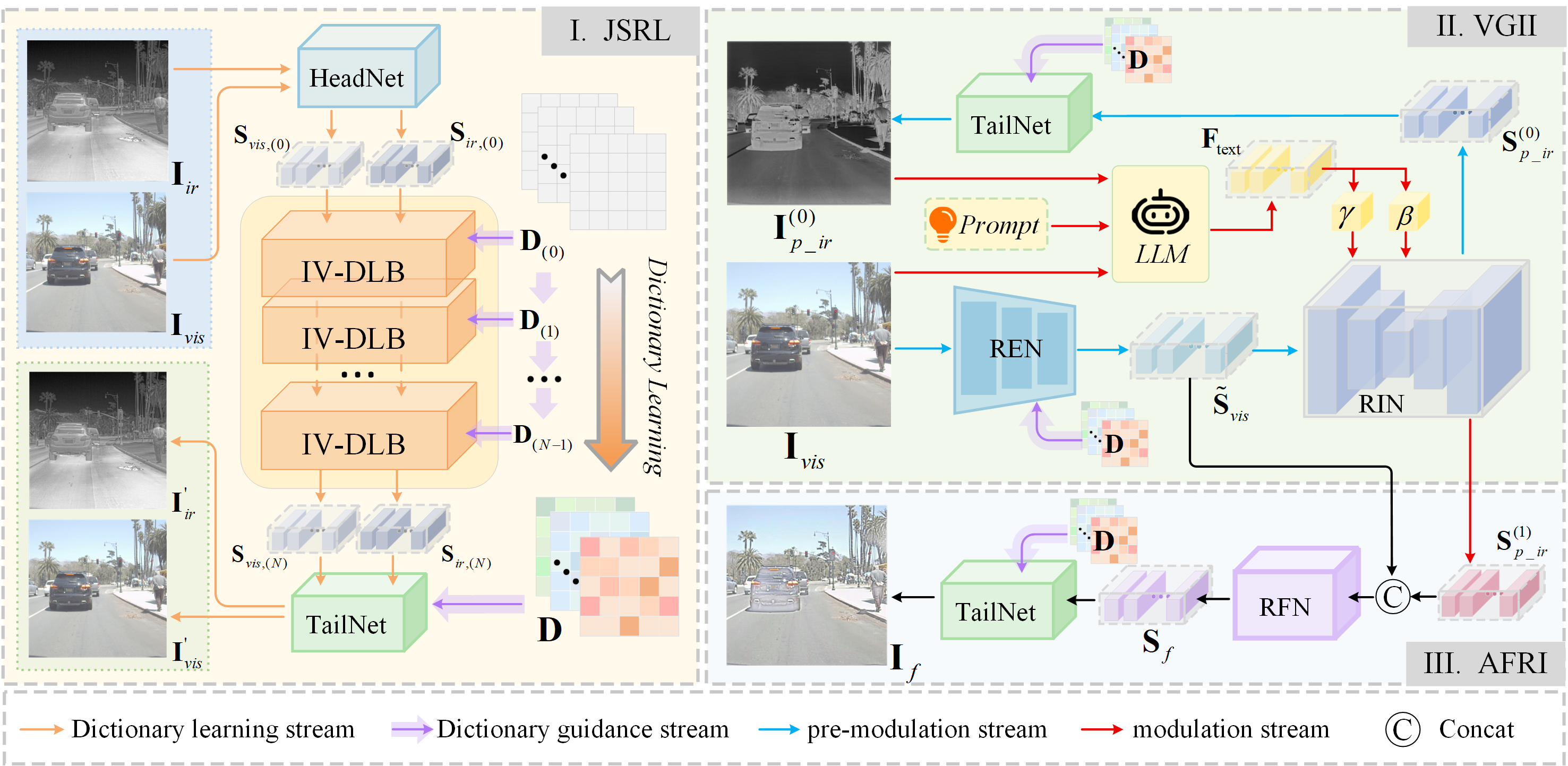}\vspace{-2mm}
	\caption{Architecture of the proposed framework. It integrates three modules (JSRL, VGII, and AFRI) to form a closed-loop pipeline of encoding, inference, fusion, and reconstruction. JSRL learns a shared dictionary for cross-modal alignment, VGII predicts latent infrared coefficients from visible ones, and AFRI fuses both modalities in the coefficient domain to reconstruct consistent, high-quality images under missing-IR conditions. }\vspace{-5mm}
	\label{fig2}
\end{figure*}

To this end, we propose a shared-dictionary, atom-correspondence forecasting framework (Figure \ref{fig2}) composed of three synergistic modules: Joint Shared-dictionary Representation Learning (JSRL), VIS-Guided IR Inference (VGII), and Adaptive Fusion via Representation Inference (AFRI). JSRL learns a cross-modal shared convolutional dictionary that projects IR and VIS onto a common atom space, thereby establishing atom-level correspondences in the dictionary-coefficient space. VGII builds a coefficient-domain transfer from VIS to IR by mapping visible coefficients to pseudo-IR coefficients and applying a lightweight refinement to enhance thermal coverage. AFRI then feeds VIS and predicted IR coefficients into an adaptive fusion network. The network outputs fused coefficients, from which the final image is reconstructed using the shared dictionary.

Compared with prior work, our method differs in three substantive ways: (i) modeling and inference are performed entirely in the coefficient domain under a shared dictionary, forming a closed loop of \emph{encode $\rightarrow$ transfer $\rightarrow$ fuse $\rightarrow$ reconstruct} instead of pixel-space black-box generation; (ii) explicit atom-level correspondences render the transfer and fusion processes interpretable; and (iii) a weak semantic prior is introduced via simple linear modulation instead of a heavy generative head, enabling controllable thermal completion with improved stability. As a result, under missing-IR conditions our approach preserves IR target saliency while maintaining VIS structural detail. \textbf{Our work makes the first attempt to explore how the image-fusion paradigm can enhance perceptual quality and downstream performance from a single visible image when the infrared modality is missing}. The main contributions are as follows:

\begin{itemize}
	\item \textbf{Dictionary-guided coefficient-domain paradigm.} 
	We propose a unified atom space in which all computations---encode, transfer, fuse, and reconstruct---are performed in the dictionary--coefficient representation, closing the loop beyond pixel space and keeping representations and constraints in the same domain to boost interpretability and robustness.
	
	\item \textbf{Controllable completion with a weak semantic prior.} 
	We introduce a frozen language model as a lightweight semantic prior that acts only as channel or atom-wise linear modulation in the coefficient domain, enabling a single, controllable correction for VIS$\rightarrow$IR transfer and reducing missed thermal cues without introducing pixel-level artifacts or instability.
	
	\item \textbf{Simple training and low-overhead inference.}
	Inference requires no real IR---only a visible image and the shared dictionary. Training uses reconstruction and simple consistency losses, avoiding adversarial or diffusion machinery. The system maintains stability while remaining reproducible, deployable, and efficient.
\end{itemize}

\section{Related Work}
\label{sec:formatting}
\subsection{Typical IR-VIS Image Fusion}
IR-VIS image fusion methods can be categorized into four groups based on their backbone architectures: CNN-based \cite{14,23,29,30,50,52} , CNN-Transformer hybrid \cite{3,5,31,40,41,43} , GAN-based \cite{24,25,26,8,42,44} , and diffusion model-based \cite{28,27,9,46,51} approaches.

CNN-based methods exploit the strength of convolution  in local pattern modeling. Through multi-scale convolution, residual aggregation, and texture enhancement, they improve detail and contrast, as in SEDRFuse \cite{29}, MLFusion \cite{23}, and CHITNet \cite{30}. These models are stable and lightweight but struggle to capture global dependencies, leading to inconsistent contrast or structural mismatch under complex illumination or cross-modal variation.
CNN-Transformer hybrids, such as IVFWSR \cite{3}, CoCoNet \cite{5} and SwinFusion \cite{31} combine convolutional locality with Transformer-based global modeling. While achieving better dependency representation, they often suffer from high parameter scale, unstable training, and reduced inference efficiency. GAN-based methods enhance perceptual realism via adversarial learning. Dual- and attention-based discriminators, such as DDcGAN \cite{24}, SDDGAN \cite{25}, and GAN-FM \cite{26}, impose multi-dimensional perceptual constraints, improving subjective quality but risking mode collapse and convergence instability. Diffusion model-based approaches enhance high-fidelity generation and cross-modal consistency by modeling the denoising process as a Markov chain. For instance, Dif-Fusion \cite{27} captures multi-channel input distributions to improve color and brightness fidelity, whereas DRMF \cite{28} adopts multi-stage denoising for balanced structure–texture restoration. Although effective for complex content, these methods entail high sampling cost and slow convergence.

Existing methods assume that both modalities are available during training and inference, and learn end-to-end mappings within the pixel or feature domain. When the IR modality is missing, these black-box generative paradigms often fail to maintain physical consistency and controllability. To address these limitations, this paper proposes an interpretable fusion framework for scenarios where the infrared modality is unavailable. 

\subsection{Model-Driven Unfolding Networks}
Model-driven unfolding transforms iterative optimization into a hierarchical framework based on the data-consistency $\rightarrow$ prior pattern. It maps iterations of classical algorithms to network layers and replaces hand-tuned hyperparameters with learnable ones. Networks thus learn step sizes, thresholds, and proximal operators in an end-to-end manner, enhancing interpretability and adaptability for complex tasks.

In dictionary learning and sparse coding, DKSVD \cite{32} replaces the $\ell_0$ prior with an $\ell_1$ prior and unfolds sparse coding via ISTA \cite{45} while parameterizing dictionaries with MLPs. Learned-CSC \cite{33,34,36} integrates convolutional sparse coding with ISTA-based networks. Multi-scale feature extraction has been incorporated before ISTA unfolding for low-level vision tasks such as artifact removal. ADMM unfolding \cite{39,35} accelerates MRI reconstruction, while joint dictionaries with dual-branch LISTA unfolding enable multi-modal super-resolution. Proximal methods \cite{37,38} for convolutional sparse models with side information enhance multi-modal performance, and CNN-embedded iterative alignment improves MRI-CT registration. Proximal-gradient networks benefit multi-hyperspectral fusion, and learnable modality-aware units improve medical image segmentation accuracy.

This paradigm injects structured priors through explicit data consistency and learnable corrections, often outperforming black-box models under missing-modality or misalignment conditions. Building on this idea, we propose a unified convolutional dictionary framework that leverages the traditional IR-VIS fusion paradigm to enhance visible-image perception and improve downstream task performance when the infrared modality is missing.

\section{Methodology}
\subsection{Overview}
To address cross-modal incompleteness caused by the absence of the infrared image, we propose a correspondence prediction-based fusion framework grounded in dictionary learning. The core idea is to predict latent infrared representations from visible ones via a shared dictionary, enabling effective fusion under missing-modality conditions. As illustrated in Figure \ref{fig2}, the framework comprises three modules: JSRL, VGII, and AFRI. JSRL learns a shared dictionary that establishes structural correspondence between the two modalities. VGII exploits structural cues in visible coefficients to infer infrared coefficients through interpretable coefficient-domain prediction rather than direct generative reconstruction. AFRI fuses visible and inferred infrared coefficients and reconstructs the final image with the shared dictionary. Through their collaborative operation, the framework enhances perceptual quality and downstream task performance under missing-infrared conditions.

\subsection{Joint Representation Learning}
We propose JSRL to learn a cross-modal \emph{shared dictionary} that represents visible and infrared images within a unified atom space.
This shared representation provides a stable and interpretable foundation for subsequent visible-guided infrared inference and fusion.
The objective jointly minimizes reconstruction errors of both modalities while imposing coefficient priors and dictionary regularization:
\begin{equation}\small
	\begin{aligned}
		\min_{\mathbf{D},\,\mathbf{S}_{vis},\,\mathbf{S}_{ir}}\;\;
		& \tfrac{1}{2}\big\|\mathbf{I}_{vis} - \mathbf{D} * \mathbf{S}_{vis}\big\|_F^2
		+ \tfrac{1}{2}\big\|\mathbf{I}_{ir} - \mathbf{D} * \mathbf{S}_{ir}\big\|_F^2 \\
		&\quad + \lambda_{1}\,\varphi_{1}\!\left(\mathbf{S}_{vis}\right)
		+ \lambda_{2}\,\varphi_{2}\!\left(\mathbf{S}_{ir}\right)
		+ \lambda_{3}\,\phi\!\left(\mathbf{D}\right),
	\end{aligned}
	\label{eq:jrl_obj}
\end{equation}
Here, $*$ denotes convolution. $\mathbf{I}_{vis}$ and $\mathbf{I}_{ir}$ represent the visible and infrared images; $\mathbf{D}$ denotes the shared dictionary; and $\mathbf{S}_{vis}$ and $\mathbf{S}_{ir}$ denote the corresponding coefficient maps. $\varphi_{1}\left(\mathbf{S}_{vis}\right)$ and $\varphi_{2}\left(\mathbf{S}_{ir}\right)$ denote priors on the coefficients (typically a sparsity prior), and $\phi\left(\mathbf{D}\right)$ denotes a dictionary regularization term, such as column-wise normalization and near-orthogonality. $\lambda_{1}$, $\lambda_{2}$ and $\lambda_{3}$ are the balance parameters.

In objective (1), if $\varphi_{1}(\mathbf{S}_{vis})$, $\varphi_{2}(\mathbf{S}_{ir})$, and $\phi(\mathbf{D})$ are predefined, the dictionary $\mathbf{D}$ and coefficients can be explicitly obtained through an optimization algorithm. However, manually defined priors and regularizations make the process complex and less generalizable to large-scale or diverse data. Inspired by \cite{1,2}, we introduce deep subnetworks, CoeNet and DicNet, which autonomously learn the joint representations $\mathbf{S}_{vis}$ and $\mathbf{S}_{ir}$ as well as the dictionary $\mathbf{D}$ from data. Built upon the theoretical foundation of objective (1), the network unifies dictionary learning and coefficient inference, embedding prior and regularization knowledge into trainable parameters. This design eliminates handcrafted constraints and enhances scalability and representation capability. To maintain consistency with the underlying optimization, the coefficient maps $\mathbf{S}_{vis}$ and $\mathbf{S}_{ir}$ output by CoeNet are designed to satisfy
\begin{equation}\small
	\begin{aligned}
		\min_{\mathbf{S}_{vis},\,\mathbf{S}_{ir}}\;\;
		& \tfrac{1}{2}\big\|\mathbf{I}_{vis} - \mathbf{D} * \mathbf{S}_{vis}\big\|_F^2
		+ \tfrac{1}{2}\big\|\mathbf{I}_{ir} - \mathbf{D} * \mathbf{S}_{ir}\big\|_F^2 \\
		&\quad + \lambda_1\,\varphi_1(\mathbf{S}_{vis})
		+ \lambda_2\,\varphi_2(\mathbf{S}_{ir}),
	\end{aligned}
	\label{eq:2}
\end{equation}

Since $\varphi_{1}(\mathbf{S}_{vis})$ and $\varphi_{2}(\mathbf{S}_{ir})$ are implicit, CoeNet cannot be directly trained using objective (\ref{eq:2}). 
To address this, auxiliary variables $\mathbf{S}'_{vis}$ and $\mathbf{S}'_{ir}$ are introduced following the method in \cite{2}, which decomposes objective (\ref{eq:2}) into
\begin{equation}
	\begin{minipage}{0.9\linewidth}
		\resizebox{\linewidth}{!}{$
			\begin{aligned}
				\min_{\mathbf{S}_{vis},\,\mathbf{S}'_{vis}}\;&
				\tfrac{1}{2}\big\|\mathbf{I}_{vis} - \mathbf{D} * \mathbf{S}'_{vis}\big\|_F^2
				+ \tfrac{\mu_1}{2}\big\|\mathbf{S}_{vis} - \mathbf{S}'_{vis}\big\|_F^2 + \lambda_1 \varphi_1(\mathbf{S}_{vis}), \\
				\min_{\mathbf{S}_{ir},\,\mathbf{S}'_{ir}}\;&
				\tfrac{1}{2}\big\|\mathbf{I}_{ir} - \mathbf{D} * \mathbf{S}'_{ir}\big\|_F^2
				+ \tfrac{\mu_2}{2}\big\|\mathbf{S}_{ir} - \mathbf{S}'_{ir}\big\|_F^2 + \lambda_2 \varphi_2(\mathbf{S}_{ir}),
			\end{aligned}
			$}
	\end{minipage}
	\label{eq:3_4}
\end{equation}
where $\mathbf{S}'_{vis}$ and $\mathbf{S}'_{ir}$ are updated in the \emph{data-consistency step}:
\begin{equation}\small
	\begin{aligned}
		\min_{\mathbf{S}'_{vis}}\;&
		\tfrac{1}{2}\big\|\mathbf{I}_{vis} - \mathbf{D} * \mathbf{S}'_{vis}\big\|_F^2
		+ \tfrac{\mu_1}{2}\big\|\mathbf{S}_{vis} - \mathbf{S}'_{vis}\big\|_F^2, \\
		\min_{\mathbf{S}'_{ir}}\;&
		\tfrac{1}{2}\big\|\mathbf{I}_{ir} - \mathbf{D} * \mathbf{S}'_{ir}\big\|_F^2
		+ \tfrac{\mu_2}{2}\big\|\mathbf{S}_{ir} - \mathbf{S}'_{ir}\big\|_F^2,
	\end{aligned}
	\label{eq:5_6}
\end{equation}
These subproblems can be solved in the Fourier domain, from which closed-form solutions can be derived using the Sherman--Morrison formula \cite{47} .
The coefficient maps $\mathbf{S}_{vis}$ and $\mathbf{S}_{ir}$ are then updated via the following proximal subproblems:
\begin{equation}\small
	\begin{aligned}
		\min_{\mathbf{S}_{vis}}\;&
		\tfrac{\beta_1}{2}\big\|\mathbf{S}_{vis} - \mathbf{S}'_{vis}\big\|_F^2
		+ \varphi_1(\mathbf{S}_{vis}), \\
		\min_{\mathbf{S}_{ir}}\;&
		\tfrac{\beta_2}{2}\big\|\mathbf{S}_{ir} - \mathbf{S}'_{ir}\big\|_F^2
		+ \varphi_2(\mathbf{S}_{ir}),
	\end{aligned}
	\label{eq:5}
\end{equation}
These proximal updates can be expressed as:
\begin{equation}\small
	\begin{aligned}
		\mathbf{S}_{vis} &= \operatorname{CoeNet}(\mathbf{S}'_{vis}, \beta_1), \quad
		\mathbf{S}_{ir} = \operatorname{CoeNet}(\mathbf{S}'_{ir}, \beta_2),
	\end{aligned}
\end{equation}
where $\mu_1$, $\mu_2$, $\beta_1$ and $\beta_2$ are adaptive implicit hyperparameters predicted by the subnetwork \textit{HypNet}.
They dynamically control the update strength according to the input content and iteration stage, enabling learnable modulation of prior influence within the unfolding framework.

Similarly, an auxiliary variable $\mathbf{D}'$ is introduced to formulate the dictionary learning problem as
\begin{equation}\small
	\begin{aligned}
		\min_{\mathbf{D},\,\mathbf{D}'}\;\;
		& \tfrac{1}{2}\big\|\mathbf{I}_{vis} - \mathbf{D}' * \mathbf{S}_{vis}\big\|_{F}^{2}
		+ \tfrac{1}{2}\big\|\mathbf{I}_{ir} - \mathbf{D}' * \mathbf{S}_{ir}\big\|_{F}^{2} \\
		& \quad + \tfrac{\mu_{3}}{2}\big\|\mathbf{D} - \mathbf{D}'\big\|_{F}^{2}
		+ \lambda_{3}\,\phi(\mathbf{D}),
	\end{aligned}
	\label{eq:9}
\end{equation}
where $\mathbf{D}'$ is updated in the \emph{data-consistency step}:
\begin{equation}\small
	\begin{aligned}
		\min_{\mathbf{D}'}\;\;
		& \tfrac{1}{2}\big\|\mathbf{I}_{vis} - \mathbf{D}' * \mathbf{S}_{vis}\big\|_{F}^{2}
		+ \tfrac{1}{2}\big\|\mathbf{I}_{ir} - \mathbf{D}' * \mathbf{S}_{ir}\big\|_{F}^{2} \\
		& \quad + \tfrac{\mu_{3}}{2}\big\|\mathbf{D} - \mathbf{D}'\big\|_{F}^{2},
	\end{aligned}
	\label{eq:10}
\end{equation}
and $\mathbf{D}$ is updated through a proximal mapping:
\begin{equation}
	\small
	\begin{minipage}{0.55\linewidth}
		\resizebox{\linewidth}{!}{$
			\begin{aligned}
				\arg\min_{\mathbf{D}}
				\tfrac{\beta_{3}}{2}\big\|\mathbf{D} - \mathbf{D}'\big\|_{F}^{2}
				+ \phi(\mathbf{D}),
			\end{aligned}
			$}
	\end{minipage}
	\label{eq:11}
\end{equation}
This can be implemented using DicNet as follows:
\begin{equation}
	\small
	\begin{minipage}{0.35\linewidth}
		\resizebox{\linewidth}{!}{$
			\begin{aligned}
				\mathbf{D} = \operatorname{DicNet}(\mathbf{D}', \beta_{3}),
			\end{aligned}
			$}
	\end{minipage}
	\label{eq:12}
\end{equation}
where $\mu_3$ and $\beta_3$ is also predicted by the \textit{HypNet}.
\begin{figure}[t!]
	\centering
	\includegraphics[width=0.425\textwidth]{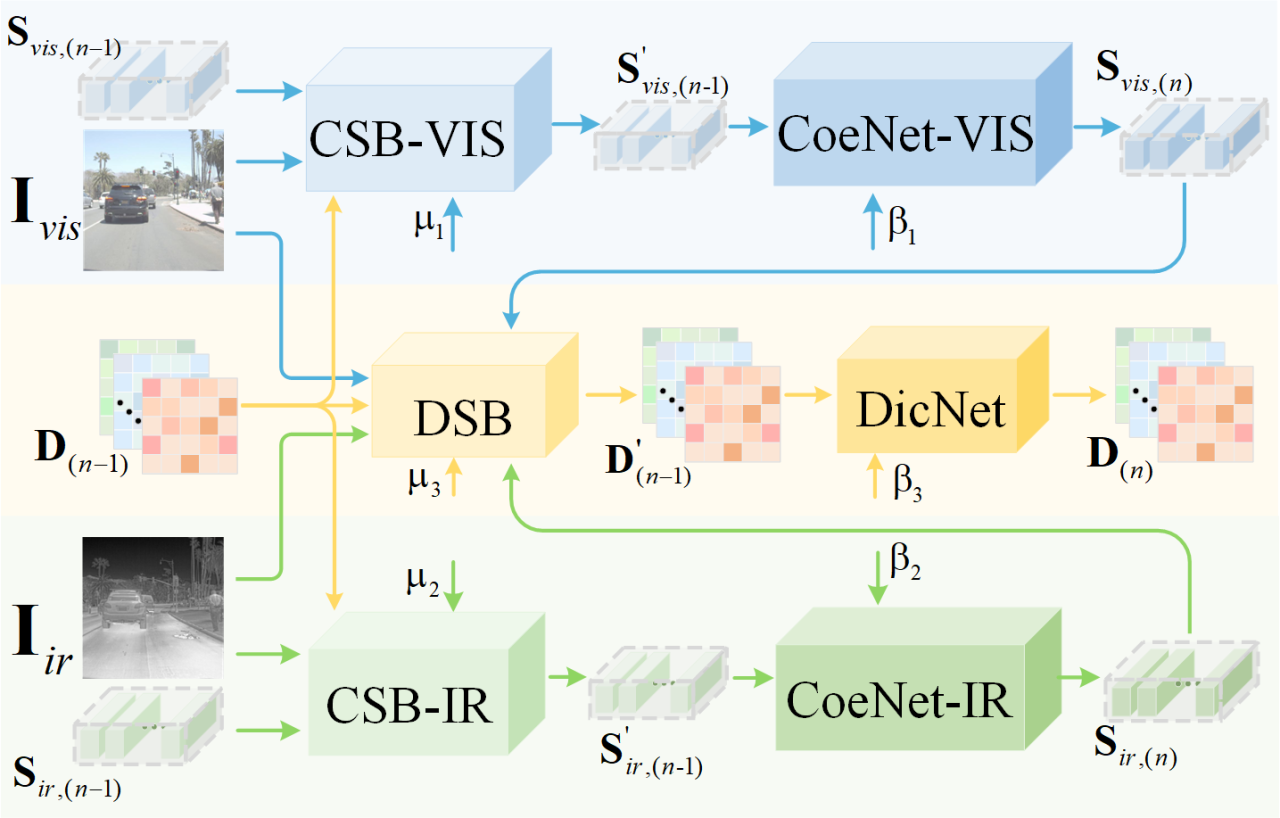}\vspace{-3mm}
	\caption{Structure of the IV-DLB.}\vspace{-4mm}
	\label{fig3}
\end{figure}

To ensure consistency between the coefficient maps $\mathbf{S}_{vis}$, $\mathbf{S}_{ir}$ predicted by CoeNet and the shared dictionary $\mathbf{D}$ predicted by DicNet in satisfying Eqs.~(\ref{eq:5}) and (\ref{eq:11}),
the networks are first trained with reconstruction-based losses:
\begin{equation}
	\small
	\begin{minipage}{0.89\linewidth}
		\resizebox{\linewidth}{!}{$
			\begin{aligned}
				\ell_S &= 
				\big\|\mathbf{D}_{(n-1)} * \mathbf{S}_{vis,(n)} - \mathbf{I}_{vis}\big\|_{1}
				+ \big\|\mathbf{D}_{(n-1)} * \mathbf{S}_{ir,(n)} - \mathbf{I}_{ir}\big\|_{1}, \\
				\ell_D &= 
				\big\|\mathbf{D}_{(n)} * \mathbf{S}_{vis,(n)} - \mathbf{I}_{vis}\big\|_{1}
				+ \big\|\mathbf{D}_{(n)} * \mathbf{S}_{ir,(n)} - \mathbf{I}_{ir}\big\|_{1},
			\end{aligned}
			$}
	\end{minipage}
	\label{eq:12_13}
\end{equation}
where $n=1,2,\cdots,N$ denotes the $n$-th update step.

As shown in Figure \ref{fig2}, the dictionary generation network consists of $N$ cascaded Infrared-Visible Dictionary Learning Blocks (IV-DLBs), with a HeadNet for input decomposition and a TailNet for image reconstruction. The HeadNet includes two convolutional layers and a ReLU. As shown in Figure \ref{fig3}, Each IV-DLB contains two coefficient solvers and a dictionary solver. The coefficient solver comprises a Coefficient Solving Block (CSB) and a CoeNet. CSB performs frequency-domain orthogonalization to reduce atom correlation, while CoeNet adopts a U-Net with symmetric residual encoder-decoder modules. The dictionary solver, consisting of a Dictionary Solving Block (DSB) and a DicNet. DSB updates kernels $\mathbf{d}$ via proximal regularized least squares and solves all spatial positions in one pass using Cholesky decomposition. DicNet stacks three residual convolutional modules with ReLU activations. TailNet uses linear convolution with reflection padding to preserve boundaries. This architecture realizes the data-consistency (Eqs.~(\ref{eq:5_6}), (\ref{eq:10})) and proximal-update (Eqs.~(\ref{eq:5}), (\ref{eq:11})) steps, embedding the shared cross-modal dictionary and interpretable atom space into trainable components. Finally, \textit{HypNet} takes the scale factor $\sigma$ as input and predicts stage-wise hyperparameters using two $1{\times}1$ convolutions, followed by a Softplus activation to ensure positivity.

\subsection{Visible-Guided Infrared Inference}
In the previous section, a unified atom space based on the shared cross-modal dictionary is established. 
Unlike direct pixel-level generation, the modality transfer from visible to infrared is performed in the dictionary-coefficient domain, ensuring structural consistency and interpretability. 
Specifically, the visible image $\mathbf{I}_{vis}$ is encoded by representation encode network (REN) as $\mathbf{\tilde{S}}_{vis} = \operatorname{REN}(\mathbf{I}_{vis};\,\mathbf{D})$, where REN includes the pre-trained \textit{HeadNet}, CSB-VIS, and CoeNet-VIS from JSRL. 
All REN parameters are frozen during coefficient generation to strictly follow the “data-consistency + proximal” principle in Eqs.~(\ref{eq:5_6}), ~(\ref{eq:5}), preserving the physical interpretability of the shared dictionary $\mathbf{D}$ and avoiding task-specific drift. 
This process projects visible information into the unified atom space, mitigating cross-modal inconsistencies common in pixel-domain learning. 
To infer the corresponding infrared coefficients, a Representation Inference Network (RIN) is introduced: 
$\mathbf{S}_{p\_ir}^{(0)} = \operatorname{RIN}(\mathbf{\tilde{S}}_{vis})$. 
RIN employs an encoder-decoder architecture with attention-based downsampling and upsampling modules, each integrating multi-head attention for spatial and channel interaction. 
By operating in the coefficient domain rather than the pixel domain, RIN directly learns atom-level relationships within the shared dictionary, naturally inheriting the interpretability and structural coherence of JSRL.

A single transfer may fail to capture fine thermal cues in real infrared images, such as subtle temperature contrasts under low illumination. 
To refine these cues, a Large Language Model (LLM) is introduced as a semantic critic. The initial pseudo-infrared image is reconstructed via the shared dictionary and TailNet:
$
\mathbf{I}_{p\_ir}^{(0)} = \operatorname{TailNet}\!(\mathbf{S}_{p\_ir}^{(0)},\,\mathbf{D}).
$
The pair $\{\mathbf{I}_{vis},\,\mathbf{I}_{p\_ir}^{(0)}\}$ and a task description form a prompt fed into the frozen LLM to extract text features $\mathbf{F}_{text}$. 
Two multi-layer perceptrons then predict Feature-wise Linear Modulation parameters:
\begin{equation}\small
	\gamma = \Phi_{\gamma}(\mathbf{F}_{text}), \qquad 
	\beta = \Phi_{\beta}(\mathbf{F}_{text}).
	\label{eq:film}
\end{equation}
Semantic-guided affine modulation is applied in the coefficient domain to yield refined visible coefficients:
\begin{equation}\small
	\mathbf{S}_{fm} = \gamma \odot \mathbf{\tilde{S}}_{vis} + \beta,
	\label{eq:sfm}
\end{equation}
which are passed through RIN for the second-stage transfer and final reconstruction:
\begin{equation}\small
	\mathbf{S}_{p\_ir}^{(1)} = \operatorname{RIN}(\mathbf{S}_{fm}), \qquad 
	\mathbf{I}_{p\_ir} = \operatorname{TailNet}\!\left(\mathbf{S}_{p\_ir}^{(1)},\,\mathbf{D}\right).
	\label{eq:srtn_tailnet}
\end{equation}

Rather than generating pixels or discriminating outputs, the LLM provides a weak semantic prior to modulate atomic responses in the coefficient space, enhancing thermal semantics without introducing generative noise. 
This closed-loop refinement operates entirely within the interpretable coefficient domain, complementing the shared dictionary to improve semantic completion under missing-modality conditions.

To ensure data consistency, thermal alignment, and structural fidelity, the total infrared inference loss is defined as
\begin{equation}\small
	\ell_{inf} = \ell_{int} + \ell_{reg} + \ell_{grad}.
	\label{eq:total_loss}
\end{equation}
The consistency term constrains the pseudo-infrared image to match the real infrared in both image and coefficient domains:
\begin{equation}\small
	\ell_{int} = 
	\|\mathbf{I}_{p\_ir} - \mathbf{I}_{ir}\|_1
	+ \|\mathbf{S}_{p\_ir}^{(1)} - \mathbf{S}_{ir}\|_1.
	\label{eq:int}
\end{equation}
Thermal regularization enforces intensity alignment using a normalized weighting map $\mathbf{A}_{ir}$:
\begin{equation}
	\small
	\begin{minipage}{0.89\linewidth}
		\resizebox{\linewidth}{!}{$
			\ell_{reg} = 
			\|\mathbf{A}_{ir} \odot \mathbf{I}_{p\_ir} - \mathbf{I}_{ir}\|_1, \quad
			\mathbf{A}_{ir} =
			\frac{\mathbf{I}_{ir} - \min(\mathbf{I}_{ir})}
			{\max(\mathbf{I}_{ir}) - \min(\mathbf{I}_{ir}) + \varepsilon}
			$}, 
	\end{minipage}
	\label{eq:reg}
\end{equation}
where $\varepsilon{=}10^{-8}$ avoids division by zero and $\odot$ denotes element-wise multiplication. 
The gradient loss anchors geometric consistency by aligning image edges:
\begin{equation}\small
	\ell_{grad} = 
	\|\nabla \mathbf{I}_{p\_ir} - \nabla \mathbf{I}_{vis}\|_1.
	\label{eq:grad}
\end{equation}

This design integrates coefficient-domain transfer, semantic modulation, and shared-dictionary reconstruction, extending the interpretable “data-consistency + proximal” paradigm of JSRL. 
By introducing a weak semantic prior, it achieves stable and reliable visible-guided infrared inference under missing-modality conditions.
\begin{figure*}[t!]
	\centering
	\includegraphics[width=1\textwidth]{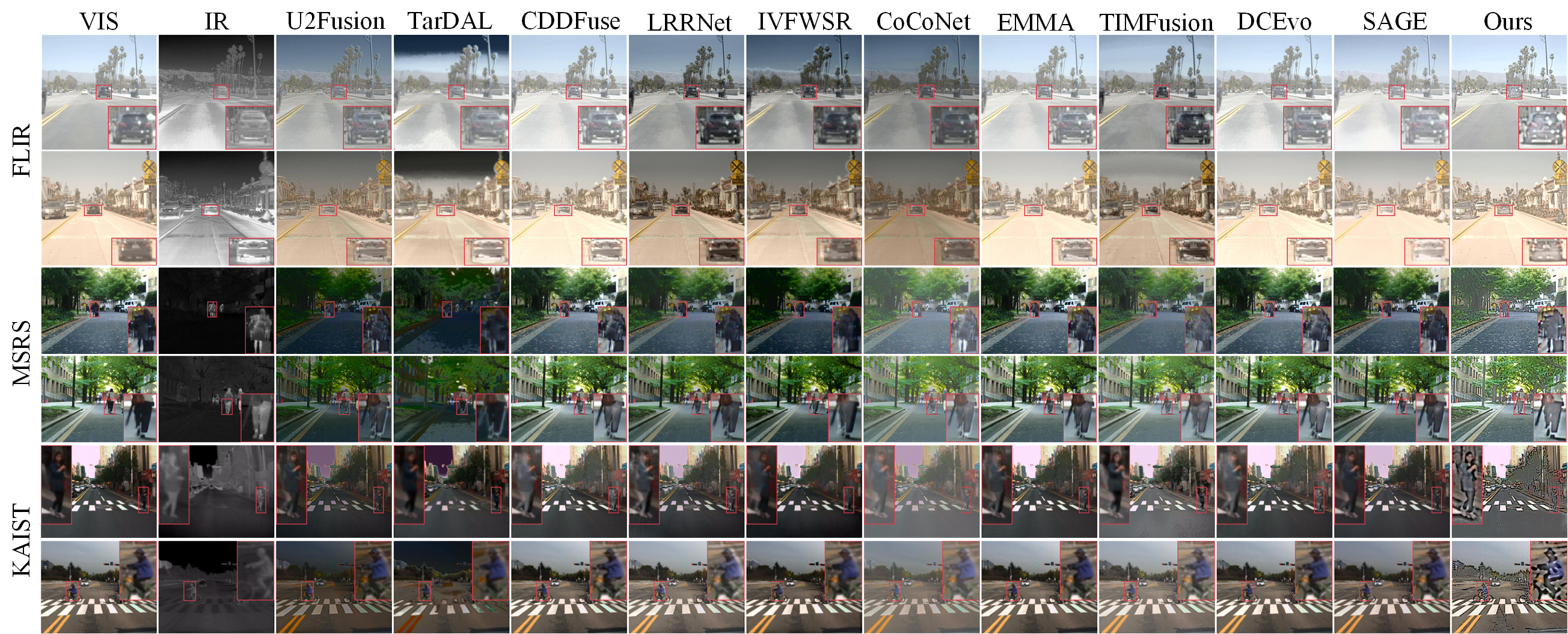}\vspace{-2mm}
	\caption{Visual comparison results on the FLIR, MSRS, and KAIST datasets for fused image quality.}\vspace{-4mm}
	\label{fig4}
\end{figure*}

\subsection{Adaptive Fusion with Representation Inference}
To adaptively integrate the structural information of the visible modality with the predicted infrared semantics in the coefficient domain, a Reasoning Fusion Network (RFN) is constructed. RFN consists of two cascaded Convolution-Attention Fusion (CAF) blocks, each combining window self-attention with convolutional mixing. 
The first block uses \{Conv, ReLU, Conv, ReLU, Conv\}, and the second \{Conv, ReLU, Conv\}, enabling local-to-global reasoning across atom and channel dimensions. Given the inputs $\mathbf{\tilde{S}}_{vis}$ and $\mathbf{S}_{p\_ir}^{(1)}$, an implicit atom-wise gating mechanism
$(\mathbf{W}_{vis},\,\mathbf{W}_{p\_ir}) =\Psi_{gate}(\operatorname{Concat}(\mathbf{\tilde{S}}_{vis},\,\mathbf{S}_{p\_ir}^{(1)}))$
learns adaptive weights for each atom, producing the fused coefficients
\begin{equation}\small
	\mathbf{S}_{f} =
	\mathbf{W}_{vis} \odot \mathbf{\tilde{S}}_{vis} +
	\mathbf{W}_{p\_ir} \odot \mathbf{S}_{p\_ir}^{(1)},
\end{equation}
where $\odot$ denotes element-wise multiplication. This gating operates in the coefficient domain, allowing each dictionary atom to adaptively draw from visible or inferred infrared cues according to saliency. Atoms along structural edges favor $\mathbf{\tilde{S}}_{vis}$, while those encoding thermal semantics rely on $\mathbf{S}_{p\_ir}^{(1)}$. Consequently, the framework achieves cross-modal collaboration within the unified atom space.

The fused coefficients, along with the shared dictionary $\mathbf{D}$ are fed into TailNet, and the generated image is given by: $\mathbf{I}_{f} = \operatorname{TailNet}(\mathbf{S}_{f},\,\mathbf{D})$. Here, the reconstruction network and dictionary are consistent with those used in encoding and inference, forming a closed loop of ``encoding $\to$ inference $\to$ fusion $\to$ reconstruction''. This ensures that the fusion result lies within the interpretable subspace spanned by the shared atoms. To highlight thermal saliency while preserving visible structures, RFN is trained using intensity and gradient consistency losses:
\begin{equation}\small
	\begin{aligned}
		\ell_{int}^{fuse} &=
		\|\mathbf{I}_{f}-\max(\mathbf{I}_{p\_ir},\,\mathbf{I}_{vis})\|_{1},\\
		\ell_{grad}^{fuse} &=
		\|\nabla \mathbf{I}_{f}-\max(\nabla \mathbf{I}_{p\_ir},\,\nabla \mathbf{I}_{vis})\|_{1},
	\end{aligned}
	\label{eq:fuse_loss}
\end{equation}
The element-wise ``$\max$'' operation encourages $\mathbf{I}_{f}$ to inherit thermal intensity peaks and sharp structural edges from infrared and visible inputs, respectively.
The total fusion loss is then
\begin{equation}\small
	\ell_{f} = \ell_{int}^{fuse} + \ell_{grad}^{fuse}.
	\label{eq:fuse_total}
\end{equation}

\newcommand{\best}[1]{\colorbox{red!15}{#1}}
\newcommand{\second}[1]{\colorbox{blue!15}{#1}}
\begin{table*}[htbp]
	\centering
	\scriptsize
	\caption{Quantitative comparison of the fusion results of our method and existing methods on the MSRS, FLIR, and KAIST datasets. The best and second-best performances are highlighted with \best{Red} and \second{Blue} backgrounds, respectively.} \vspace{-2mm}
	\renewcommand{\arraystretch}{1.2}
	\setlength{\tabcolsep}{2pt}  
	\resizebox{0.95\textwidth}{!}{
		\begin{tabular}{c|cccccc|cccccc|cccccc}
			\toprule
			\multirow{2}*{\textbf{Methods}}  & \multicolumn{6}{c|}{\textbf{MSRS}} & \multicolumn{6}{c|}{\textbf{FLIR}} & \multicolumn{6}{c}{\textbf{KAIST}}\\
			\cline{2-19}
			& AG $\uparrow$ & CE $\downarrow$ & EI $\uparrow$ & EN $\uparrow$ & Qcb $\uparrow$ & SF $\uparrow$ & AG $\uparrow$ & CE $\downarrow$ & EI $\uparrow$ & EN $\uparrow$ & Qcb $\uparrow$ & SF $\uparrow$ & AG $\uparrow$ & CE $\downarrow$ & EI $\uparrow$ & EN $\uparrow$ & Qcb $\uparrow$ & SF $\uparrow$ \\
			\midrule
			U2Fusion & 2.308 & \second{0.840} & 24.692 & 6.145 & 0.483 & 6.288 & 4.645 & 1.353 & 48.215 & 6.916 & 0.410 & 11.302 & \second{3.262} & 1.626 & \second{34.797} & 6.092 & 0.505 & 9.399 \\
			TarDAL   & 2.189 & 2.183 & 23.251 & 6.059 & 0.417 & 6.379 & 3.722 & 0.643 & 39.761 & \best{7.393} & 0.420 & 9.392 & 1.815 & 1.273 & 19.393 & 6.282 & 0.431 & 5.355 \\
			CDDFuse & 4.818 & 1.609 & 51.368 & \second{7.321} & \second{0.565} & 14.250 & \best{5.079} & 0.726 & \best{52.633} & 6.766 & 0.454 & \best{13.346} & 3.167 & 1.298 & 34.043 & \best{7.122} & 0.497 & 9.194 \\
			LRRNet  & 3.621 & 2.837 & 38.519 & 6.930 & 0.461 & 10.731 & \second{4.648} & 1.798 & \second{49.145} & 7.063 & \second{0.481} & 12.449 & 2.939 & 1.475 & 31.707 & 7.032 & 0.463 & 8.213 \\
			IVFWSR  & 3.360 & 1.337 & 36.254 & 6.798 & 0.457 & 9.241 & 3.386 & 1.266 & 36.392 & \second{7.365} & 0.400 & 8.144 & 2.672 & \second{0.868} & 28.868 & 6.959 & 0.443 & 7.743 \\
			CoCoNet & 3.299 & \best{0.758} & 34.565 & 6.510 & 0.427 & 9.804 & 2.856 & 1.848 & 29.542 & 6.591 & \best{0.492} & 7.340 & 1.995 & 0.963 & 21.090 & 6.233 & 0.474 & 5.912 \\
			EMMA    & \second{4.913} & 1.343 & \second{52.642} & \best{7.333} & 0.539 & 14.176 & 3.796 & 0.745 & 40.356 & 6.489 & 0.451 & 10.188 & 3.083 & \best{0.859} & 33.405 & 7.005 & 0.514 & 8.837 \\
			TIMFusion & 3.649 & 3.796 & 38.972 & 6.846 & 0.394 & 10.771 & 4.202 & 1.541 & 44.643 & 7.030 & 0.415 & 11.768 & 3.194 & 1.899 & 34.409 & 6.998 & 0.438 & 8.831 \\
			DCEvo   & 4.858 & 1.696 & 51.810 & 7.298 & \best{0.571} & \second{14.262} & 4.585 & 0.812 & 48.435 & 6.763 & 0.465 & 12.097 & 3.229 & 1.453 & 34.459 & 7.027 & \second{0.517} & \second{9.879} \\
			SAGE    & 4.452 & 2.058 & 47.134 & 7.030 & 0.529 & 13.285 & 3.254 & \second{0.697} & 33.860 & 6.646 & 0.376 & 9.115 & 2.950 & 1.037 & 31.467 & 6.799 & \best{0.520} & 8.835 \\
			Ours    & \best{5.037} & 3.142 & \best{53.168} & 7.188 & 0.454 & \best{14.898} & 4.518 & \best{0.596} & 48.784 & 6.639 & 0.435 & \second{12.554} & \best{4.414} & 2.227 & \best{37.115} & \second{7.073} & 0.405 & \best{11.031} \\
			\bottomrule  
		\end{tabular}   
		\label{table1}
	}
\end{table*}

\section{Experiment}
\subsection{Experimental Setup}
\textbf{Datasets.} We evaluate fusion performance on the FLIR, KAIST~\cite{11}, and MSRS~\cite{12} datasets, and assess downstream performance, including object detection and semantic segmentation, on the $\text{M}^3$FD~\cite{13} and FMB~\cite{14} datasets. Specifically, we randomly select 5,445 training pairs and 765 testing pairs from FLIR, 8,310 training pairs and 1,246 testing pairs from KAIST, and 536 training pairs and 179 testing pairs from MSRS. All images are randomly cropped into $128\times128$ patches.
\begin{figure*}[t!]
	\centering
	\includegraphics[width=1\textwidth]{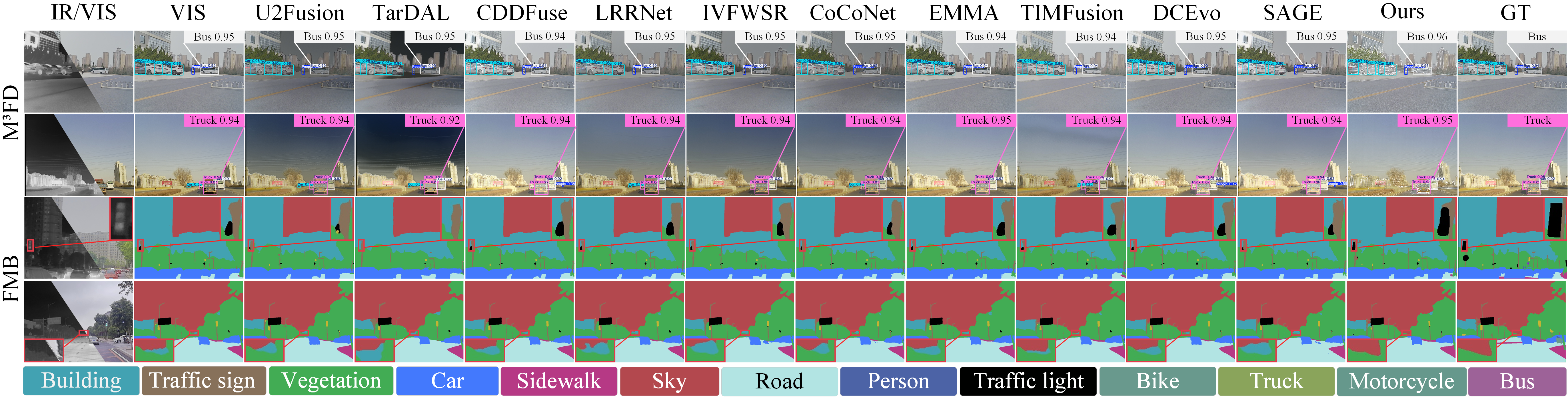}\vspace{-2mm}
	\caption{Qualitative comparison of object detection and semantic segmentation on the $\text{M}^3$FD and FMB datasets, respectively.}\vspace{-3mm}
	\label{fig5}
\end{figure*}

\textbf{Implementation Details.}  The JSRL, VGII and AFRI modules are trained sequentially. The JSRL module is trained on the MSRS for 1,000 epochs with a batch size of 24 and an initial learning rate of 0.0001. The dictionary trained by MSRS can be universally applied to other datasets in subsequent experiments. The VGII and AFRI modules are trained for 10 epochs, both with a batch size of 10 and an initial learning rate of 0.001. The Adam optimizer \cite{49} is used to update all parameters. In the JSRL module, the convolutional kernel size for the joint dictionary is set to $5\times5$, and an IV-DLB is used. All experiments are conducted on two NVIDIA GeForce RTX 4090 GPUs, and the model is implemented using the PyTorch 3.10.18 framework.

\subsection{Comparison with State-of-the-arts}
To evaluate the effectiveness of our method, we compare it with ten state-of-the-art infrared and visible image fusion methods, including U2Fusion~\cite{15}, TarDAL~\cite{13}, CDDFuse~\cite{16}, LRRNet~\cite{17}, IVFWSR~\cite{3}, CoCoNet~\cite{5}, EMMA~\cite{18}, TIMFusion~\cite{19}, DCEvo~\cite{20}, and SAGE~\cite{21}. In our experiments, the proposed method requires only visible images as input, whereas all comparison methods require both infrared and visible images.

Figure~\ref{fig4} presents the qualitative results of different methods. Our method shows superior detail fidelity, thermal information representation, brightness and contrast balance, and overall visual consistency. These observations indicate that, even without access to the infrared modality, our method can effectively recover infrared information. The quantitative comparisons are summarized in Table~\ref{table1}. The results show that our method achieves performance comparable to fusion approaches that take both infrared and visible images as input across all evaluation metrics. \textbf{It is worth noting that the supplementary material also includes comparisons with results obtained by fusing visible images with generated infrared images}.
\begin{table*}[htbp]
	\centering
	\scriptsize
	\caption{Quantitative comparison of the object detection and semantic segmentation results of our method and existing methods on the $\text{M}^3$FD and FMB datasets, respectively. Best results are highlighted in \best{\textbf{red}} and second-best in \second{\textbf{blue}}.}\vspace{-2mm}
	\renewcommand{\arraystretch}{1.2}
	\setlength{\tabcolsep}{4pt}
	\begin{tabular}{c|ccccccc|ccccccc}
		\toprule
		\multirow{2}*{\textbf{Methods}}  & \multicolumn{7}{c|}{\textbf{$\text{M}^3$FD}} & \multicolumn{7}{c}{\textbf{FMB}} \\
		\cline{2-15}
		& people & car & bus & motor & truck & lamp & mAP & wall & building & sky & floor & tree & road & mIoU \\
		\midrule
		VIS & 0.864 & 0.964 & 0.990 & 0.926 & 0.908 & 0.970 & 0.931 & 89.461 & 58.723 & 85.666 & 33.918 & 67.091 & 95.984 & 61.794 \\
		U2Fusion & 0.841 & 0.959 & 0.989 & 0.918 & 0.878 & 0.970 & 0.926 & 89.523 & 57.843 & 85.234 & 33.852 & 65.076 & 94.965 & 60.829 \\
		TarDAL & 0.896 & 0.976 & \colorbox{blue!15}{0.993} & 0.951 & 0.913 & 0.983 & 0.952 & 87.462 & 52.695 & 77.013 & 31.397 & 53.117 & 90.756 & 56.194 \\
		CDDFuse & 0.883 & 0.974 & 0.991 & 0.943 & 0.903 & 0.976 & 0.945 & 89.811 & 57.901 & \colorbox{blue!15}{87.383} & 32.988 & 67.748 & \colorbox{blue!15}{96.055} & 61.649 \\
		LRRNet & 0.898 & \colorbox{red!15}{0.978} & \colorbox{blue!15}{0.993} & 0.958 & 0.904 & \colorbox{blue!15}{0.986} & 0.953 & 89.705 & \colorbox{blue!15}{59.081} & 86.685 & \colorbox{blue!15}{34.596} & 66.974 & 95.678 & \colorbox{red!15}{62.942} \\
		IVFWSR & 0.876 & 0.974 & 0.991 & 0.948 & \colorbox{red!15}{0.927} & 0.984 & 0.952 & 88.673 & 59.003 & 85.711 & 34.307 & 63.281 & 95.324 & 60.801 \\
		CoCoNet & 0.893 & 0.974 & 0.991 & 0.947 & 0.899 & 0.982 & 0.948 & 89.630 & 57.803 & 86.597 & 32.492 & 64.982 & 95.653 & 58.513 \\
		EMMA & 0.898 & \colorbox{red!15}{0.978} & \colorbox{red!15}{0.994} & 0.956 & 0.909 & 0.984 & 0.953 & \colorbox{red!15}{90.132} & 55.686 & \colorbox{red!15}{87.584} & 33.191 & \colorbox{blue!15}{68.315} & \colorbox{blue!15}{96.055} & 61.952 \\
		TIMFusion & 0.896 & \colorbox{blue!15}{0.977} & \colorbox{blue!15}{0.993} & 0.958 & 0.913 & 0.984 & \colorbox{blue!15}{0.954} & \colorbox{blue!15}{89.911} & 56.663 & 87.015 & 32.953 & 66.018 & 95.874 & 62.836 \\
		DCEvo & 0.885 & 0.975 & 0.989 & 0.946 & 0.907 & 0.977 & 0.946 & 89.372 & 56.562 & 87.231 & 33.876 & \colorbox{red!15}{68.513} & \colorbox{red!15}{96.075} & 60.366 \\
		SAGE & \colorbox{blue!15}{0.900} & \colorbox{blue!15}{0.977} & \colorbox{blue!15}{0.993} & \colorbox{blue!15}{0.960} & \colorbox{blue!15}{0.924} & 0.985 & \colorbox{red!15}{0.956} & 88.189 & 55.754 & 86.682 & 33.715 & 66.070 & 95.774 & 60.777 \\
		Ours & \colorbox{red!15}{0.902} & 0.972 & \colorbox{red!15}{0.994} & \colorbox{red!15}{0.967} & 0.919 & \colorbox{red!15}{0.990} & 0.948 & 89.549 & \colorbox{red!15}{65.923} & 87.178 & \colorbox{red!15}{36.346} & 67.578 & 95.967 & \colorbox{blue!15}{62.939} \\
		\bottomrule
	\end{tabular}
	\label{table2}\vspace{-2mm}
\end{table*} 

\subsection{Performance on Downstream Tasks}
To further evaluate the effect of different methods on downstream tasks, we conduct object detection and semantic segmentation experiments on the $\text{M}^3$FD and FMB datasets, respectively. The results are shown in Table~\ref{table2} and Figure~\ref{fig5}. For object detection, the fused images are used to train and test the YOLOv5s\footnote{https://github.com/ultralytics/yolov5} model, while for semantic segmentation, they are used to train and test the SegFormer-b5~\cite{22} model.

As shown in Table \ref{table2}, our method exhibits performance comparable to that of full-modal fusion methods in terms of the mAP metric, with the AP values for most object categories reaching high levels. The visual results in Figure \ref{fig5} further confirm that, even in the absence of the infrared modality, the detection model is able to accurately recognize and localize objects based on the images generated by our method. The detection confidence is consistent with that of the full-modal fusion images.  Furthermore, as shown in Table \ref{table2}, in the semantic segmentation task, the performance of our method is comparable to that of full-modal fusion methods, and in some metrics, it even outperforms them. The qualitative results in Figure \ref{fig5} further illustrate that, compared to full-modal fusion methods, the images generated by our method are more accurately segmented into distinct semantic regions, with clearer boundaries and improved internal consistency. These findings strongly suggest that the high-quality images produced by our method offer enhanced adaptability for downstream tasks. 

\vspace{-3mm}
\begin{table}[htbp]
	\centering
	\scriptsize
	\caption{Quantitative comparison between our model and its ablated variants. The best result for each metric is highlighted in \best{Red}.}\vspace{-2mm}
	\label{tab:m3fd_fmb_result}
	\renewcommand{\arraystretch}{1.2}
	\setlength{\tabcolsep}{3pt}
	\resizebox{\linewidth}{!}{   
		\begin{tabular}{c|cc|cccccc}
			\toprule
			\textbf{Methods} & \textbf{Dictionary} & \textbf{LLM} & AG $\uparrow$ & CE $\downarrow$ & EI $\uparrow$ & EN $\uparrow$ & Qcb $\uparrow$ & SF $\uparrow$  \\
			\midrule
			Model $\mathrm{I}$ & \ding{55} & \ding{55} & 3.320  & 1.452  & 45.531  & 6.058  & 0.396  & 9.238 \\
			Model $\mathrm{II}$ & \ding{51} & \ding{55} & 4.363  & 1.046  & 48.351  & 6.578  & 0.434  & 11.936 \\
			Model $\mathrm{III}$ & \ding{55} & \ding{51} & 4.256  & 0.619  & 48.154  & 6.423  & 0.418  & 11.175 \\
			Ours & \ding{51} & \ding{51} & \best{4.518} & \best{0.596} & \best{48.784} & \best{6.639} & \best{0.435} & \best{12.554} \\
			\bottomrule
		\end{tabular}
	}\vspace{-4mm}
	\label{table3}
\end{table}
\begin{figure}[htbp]
	\centering
	\includegraphics[width=0.43\textwidth]{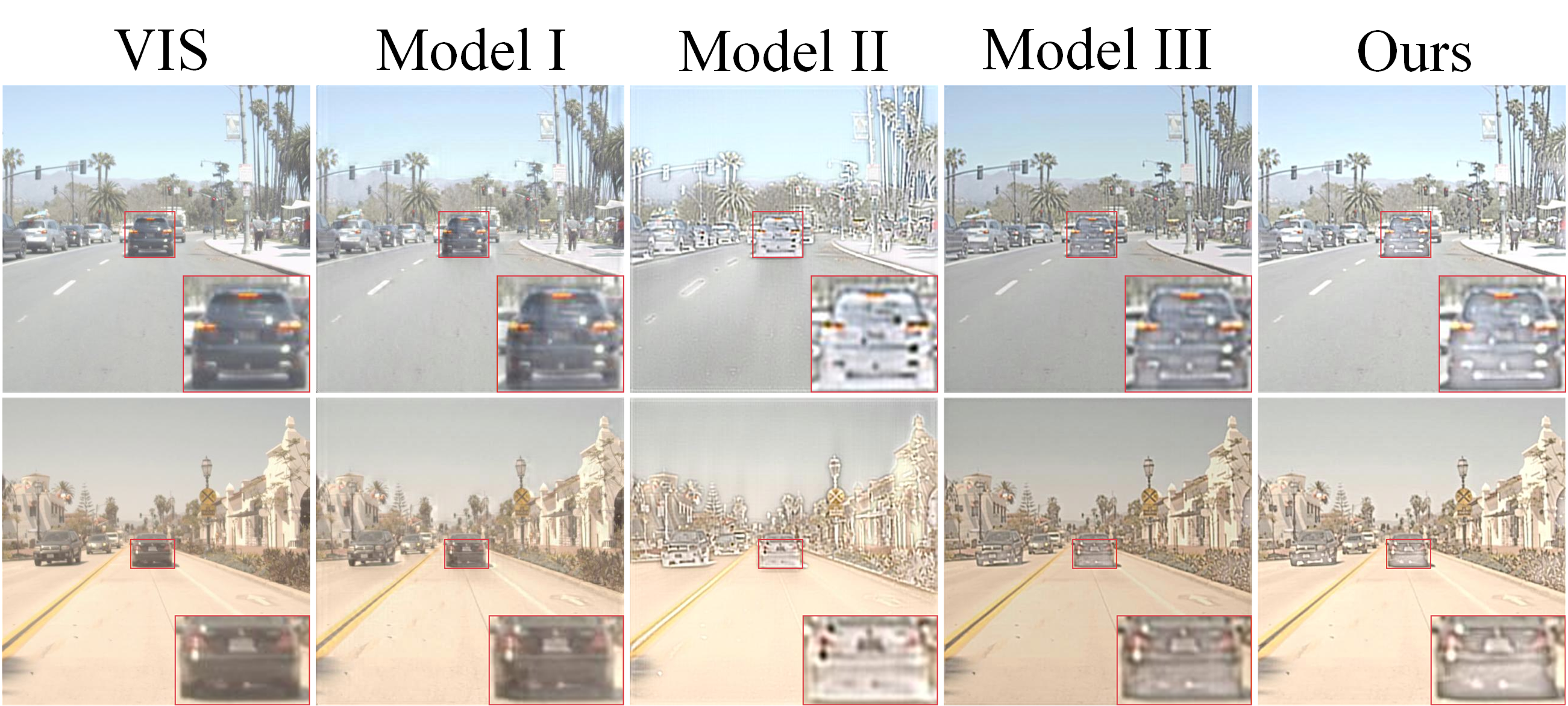}\vspace{-2mm}
	\caption{Qualitative comparison between our model and its ablated variants.}\vspace{-4mm}
	\label{fig6} 
\end{figure}

\subsection{Ablation Study}
The core of our method lies in the joint sparse dictionary and the LLM-based modulation. To evaluate the contribution of these two components to infrared and visible image fusion under the condition of missing infrared modality, we perform an ablation study on the FLIR dataset. The results are reported in Table~\ref{table3} and Figure~\ref{fig6}. We construct Model I as the baseline. In Model I, the joint dictionary is replaced with an encoder-decoder architecture that consists of three $3\times3$ convolutional layers, and the LLM-based modulation for infrared prediction is removed. As shown in Figure~\ref{fig6}, Model I fails to recover infrared information and produces blurry results with noticeable artifacts compared to the source visible images.

\textbf{Effectiveness of the Joint Sparse Dictionary.} Building upon Model I, we replace the encoder and decoder structure composed of convolutional layers with a joint dictionary, resulting in Model II. The joint dictionary guides the process of image encoding and reconstruction. As shown in Figure~\ref{fig6}, the introduction of the joint dictionary enables Model II to integrate infrared information into the generated images, thereby enhancing the thermal regions. The metrics in Table~\ref{table3} further demonstrate improved image quality. These results indicate that the joint dictionary allows the model to more effectively capture cross-modal correspondences in the joint representation space, facilitating better thermal compensation and fusion.

\textbf{Effectiveness of LLM Modulation.} Additionally, we integrate LLM guidance into Model I to provide discriminative supervision for infrared reasoning, resulting in Model III. As illustrated in Figure~\ref{fig6}, the images generated by Model III display enhanced brightness and contrast, along with clearer edge details. The quantitative results in Table~\ref{table3} further show that Model III outperforms Model I across all evaluation metrics. These findings imply that LLM effectively provides semantic constraints and discriminative cues during infrared reasoning, enabling the model to more accurately capture and preserve key thermal features.

In our full model, both the joint dictionary and LLM modulation are incorporated. As shown in Figure~\ref{fig6}, the images generated by our model achieve the best visual performance, featuring superior detail preservation, texture quality, color fidelity, and enhancement of thermal regions. Moreover, as presented in Table~\ref{table3}, our method attains the highest scores across all evaluation metrics. These results underscore the complementary contributions of the joint dictionary and LLM modulation. \textbf{More experiments and analysis are provided in the supplementary material.}

\section{Conclusion}
We propose an IR-VIS fusion method tailored for scenarios where the infrared modality is absent. This method conducts coefficient domain fusion by imposing transfer and consistency constraints between visible coefficients and pseudo infrared coefficients within a shared dictionary space. The JSRL module establishes cross-modal coefficient consistency. Under this consistency, the RIN performs interpretable reasoning for the missing modality and reconstructs thermal information from the visible representation. The AFRI module further enhances semantic consistent fusion and reconstruction by utilizing both visible and pseudo infrared representations, generating high quality images that preserve structural details and thermal cues. Experiments demonstrate that, despite the lack of infrared input, the proposed method achieves performance comparable to that of full modal fusion approaches. Moreover, it exhibits stable and superior performance in downstream tasks such as object detection and semantic segmentation.

\section*{Acknowledgments}
This work was supported in part by the National Natural Science Foundation of China(62571222, 62276120, 62576163, 62161015), and the Yunnan Fundamental Research Projects (202501AS070123, 202301AV070004).



{
    \small
    \bibliographystyle{ieeenat_fullname}
    \bibliography{main}
}

\clearpage
\appendix
\setcounter{page}{1}
\maketitlesupplementary

\section{More Details of Our Method}
We present the process of Joint Shared-dictionary Representation Learning (JSRL) in Algorithm \ref{JSRL}. First, the infrared and visible images are fed into the HeadNet to generate their coefficient maps, with the shared dictionary being initialized to zeros. Subsequently, these initial representations are progressively refined through N IV-DLBs, enabling the final coefficients and shared dictionary to accurately reconstruct the input infrared and visible images.

\begin{algorithm}[htbp]
	\small 
	\caption{Joint Shared-dictionary Representation Learning}
	\label{JSRL}
	\begin{algorithmic}[1]
		\INPUT Infrared image $\mathbf{I}_{ir}$, visible image $\mathbf{I}_{vis}$, scale factor $\sigma$, the number of IV-DLBs $N$
		\OUTPUT Reconstructed infrared image $\mathbf{I}'_{ir}$, Reconstructed visible image $\mathbf{I}'_{vis}$, shared dictionary $\mathbf{D}$
		
		\STATE $\mathbf{S}_{vis,(0)} = \text{HeadNet}(\mathbf{I}_{vis})$, $\mathbf{S}_{ir,(0)} = \text{HeadNet}(\mathbf{I}_{ir})$,
		\quad $\mathbf{D}_{(0)} = 0$
		\FOR{$n = 1$ to $N$}
		\STATE $\{\mu_{1,(n)}, \mu_{2,(n)}, \mu_{3,(n)},\beta_{1,(n)}, \beta_{2,(n)}, \beta_{3,(n)}\} = \text{HypNet}(\sigma)$
		\STATE $\mathbf{S}'_{vis,(n)} = \text{CSB-VIS}\big(\mathbf{I}_{vis}, \mathbf{D}_{(n-1)}, \mathbf{S}_{vis,(n-1)}, \mu_{1,(n)}\big)$
		\STATE $\mathbf{S}_{vis,(n)} = \text{CoeNet-VIS}\big(\mathbf{S}'_{vis,(n)}, \beta_{1,(n)}\big)$
		\STATE $\mathbf{S}'_{ir,(n)} = \text{CSB-IR}\big(\mathbf{I}_{ir}, \mathbf{D}_{(n-1)}, \mathbf{S}_{ir,(n-1)}, \mu_{2,(n)}\big)$
		\STATE $\mathbf{S}_{ir,(n)} = \text{CoeNet-IR}\big(\mathbf{S}'_{ir,(n)}, \beta_{2,(n)}\big)$
		\STATE $\mathbf{D}'_{(n)} = \text{DSB}\big(\mathbf{I}_{vis}, \mathbf{I}_{ir}, \mathbf{D}_{(n-1)}, \mathbf{S}_{vis,(n)}, \mathbf{S}_{ir,(n)}, \mu_{3,(n)}\big)$
		\STATE $\mathbf{D}_{(n)} = \text{DicNet}\big(\mathbf{D}'_{(n)}, \beta_{3,(n)}\big)$
		\ENDFOR
		\STATE $\mathbf{I}'_{vis} = \mathbf{D}_{(N)} \ast \mathbf{S}_{vis,(N)}$, $\mathbf{I}'_{ir} = \mathbf{D}_{(N)} \ast \mathbf{S}_{ir,(N)}$
	\end{algorithmic}
\end{algorithm}

In addition, we summarize the training pipeline for the VIS-Guided IR Inference (VGII) and Adaptive Fusion via Representation Inference (AFRI) modules in Algorithm \ref{AFRI}.

\begin{algorithm}[htbp]
	\small 
	\caption{VIS-Guided IR Inference and Adaptive Fusion via Representation Inference}
	\label{AFRI}
	\begin{algorithmic}[1]
		\INPUT visible image $\mathbf{I}_{vis}$, shared dictionary $\mathbf{D}$, pretrained LLM: Qwen-7B-Chat
		\OUTPUT Fusion image $\mathbf{I}_{f}$
		
		\STATE $\mathbf{\tilde{S}}_{vis} = \text{REN}(\mathbf{I}_{vis}, \mathbf{D})$
		\STATE $\mathbf{S}_{p\_ir}^{(0)} = \operatorname{RIN}(\mathbf{\tilde{S}}_{vis})$
		\STATE $\mathbf{I}_{p\_ir}^{(0)} = \operatorname{TailNet}\!(\mathbf{S}_{p\_ir}^{(0)},\,\mathbf{D})$
		\STATE $\mathbf{F}_{text} = \text{LLM}\big(\mathbf{I}_{vis},\,\mathbf{I}_{p\_ir}^{(0)})$
		\STATE $\gamma = \Phi_{\gamma}(\mathbf{F}_{text}), \qquad 
		\beta = \Phi_{\beta}(\mathbf{F}_{text})$
		\STATE $\mathbf{S}_{p\_ir}^{(1)} = \operatorname{RIN}(\gamma \odot \mathbf{\tilde{S}}_{vis} + \beta)$
		\STATE $\{\mathbf{W}_{vis},\,\mathbf{W}_{p\_ir}\} =\Psi_{gate}(\operatorname{Concat}(\mathbf{\tilde{S}}_{vis},\,\mathbf{S}_{p\_ir}^{(1)}))$
		\STATE $\mathbf{I}_{f} = \text{TailNet}(
		\mathbf{W}_{vis}, \mathbf{\tilde{S}}_{vis},
		\mathbf{W}_{p\_ir}, \mathbf{S}_{p\_ir}^{(1)})$
	\end{algorithmic}
\end{algorithm}

\section{Hyperparameter Analysis}
The proposed method involves two key hyperparameters: the kernel size of the dictionary and the number of  IV-DLBs. In our implementation, we set the kernel size of the dictionary to $5\times5$ and use a single IV-DLB layer. In this section, we analyze the influence of the hyperparameters on model performance on the FLIR dataset.

\begin{table}[htbp]
	\centering
	\scriptsize
	\caption{Influence of the dictionary kernel size on model perfomance. The best performance for each metric is marked with \best{Red}.}
	\label{tab1}
	\renewcommand{\arraystretch}{1.2}
	\setlength{\tabcolsep}{3pt}
	\resizebox{\linewidth}{!}{   
		\begin{tabular}{c|cccccc}
			\toprule
			\textbf{Kernel size} & AG $\uparrow$ & CE $\downarrow$ & EI $\uparrow$ & EN $\uparrow$ & Qcb $\uparrow$ & SF $\uparrow$ \\
			\midrule
			$3\times3$  & 4.156  & 1.357  & 44.882  & 6.545  & 0.430  & 11.195 \\
			$5\times5$  & \best{4.518}  & \best{0.596}  & \best{48.784}  & \best{6.639}  & \best{0.435}  & \best{12.554} \\
			$7\times7$  & 4.499  & 1.298  & 48.560  & 6.627  & 0.356  & 12.300 \\
			\bottomrule
		\end{tabular}
	}
\end{table}

\textbf{Influence of the dictionary kernel size.} To evaluate the impact of the dictionary kernel size on model perfomance, we evaluate three different kernel configurations: $3\times3$, $5\times5$, and $7\times7$. As shown in Table \ref{tab1}, increasing the kernel size from $3\times3$ to $5\times5$ results in notable improvements across all metrics. However, further enlarging the kernel to $7\times7$ leads to varying degrees of performance degradation. $5\times5$ yields the best overall fusion performance.

\textbf{Influence of the number of IV-DLB.} To evaluate the effect of the number of IV-DLBs, we conduct experiments with $N = 1, 2, 3$. As shown in Table \ref{tab2}, increasing $N$ from 1 to 2 leads to only marginal performance gains, while the FLOPs nearly doubles. Further increasing N to 3 results in a degradation of performance.  Therefore, we introduce an IV-DLB layer in our experiments to achieve a favorable trade-off between model performance and computational complexity.

\begin{table}[htbp]
	\centering
	\scriptsize
	\caption{Influence of the number of IV-DLBs on model perfomance. The best performance for each metric is marked with \best{Red}.}
	\label{tab2}
	\renewcommand{\arraystretch}{1.2}
	\setlength{\tabcolsep}{3pt}
	\resizebox{\linewidth}{!}{   
		\begin{tabular}{c|cccccc|c}
			\toprule
			\textbf{N} & AG $\uparrow$ & CE $\downarrow$ & EI $\uparrow$ & EN $\uparrow$ & Qcb $\uparrow$ & SF $\uparrow$ & GFLOPS \\
			\midrule
			1  & 4.518  & 0.596  & 48.784  & 6.639  & \best{0.435}  & 12.554  & \best{63.023} \\
			2  & \best{4.529}  & \best{0.565}  & \best{49.096}  & 6.598  & 0.432  & \best{12.671}  & 120.795 \\
			3  & 4.485  & 0.679  & 48.641  & \best{6.705}  & 0.432  & 12.265  & 178.567 \\
			\bottomrule
		\end{tabular}
	}
\end{table}

\begin{figure}[h]
	\centering
	\includegraphics[width=\linewidth]{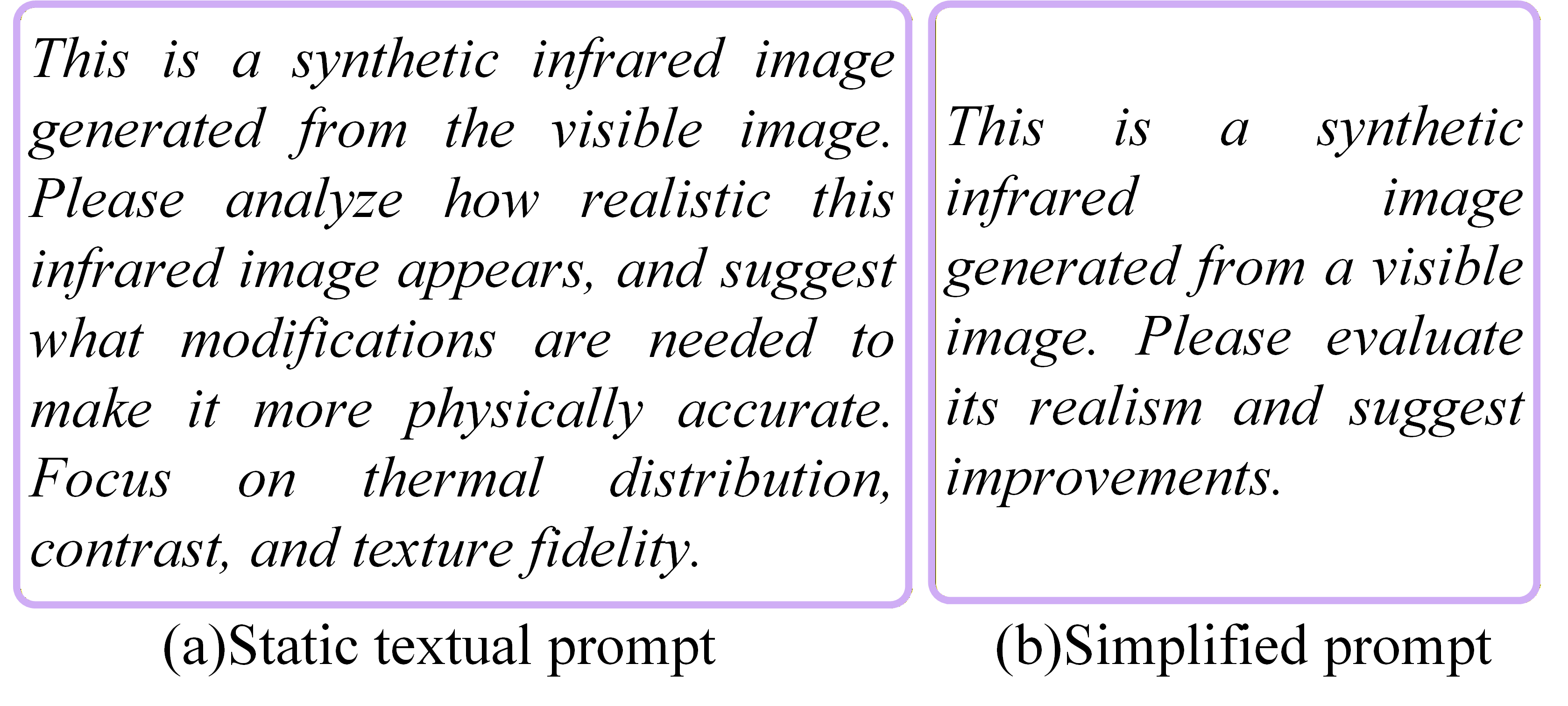}
	\caption{Two different types of text prompts. (a) is our prompt used in the experiment, while (b) is the simplified version.}
	\label{f1}
\end{figure}

\textbf{Influence of the textual prompt.} To evaluate the impact of textual prompt fed to the LLM on the model performance, we conduct experiments using three different types of prompt settings. As shown in the Table \ref{tab3}, the results demonstrate that variations in the prompts affect the fusion performance. The model utilizing the complete prompt achieves the best results. Both our prompt and a simplified version are illustrated in Figure \ref{f1}. 

\begin{table}[htbp]
	\centering
	\scriptsize
	\caption{Influence of the textual prompt on model perfomance. The best performance for each metric is marked with \best{Red}.}
	\label{tab3}
	\renewcommand{\arraystretch}{1.2}
	\setlength{\tabcolsep}{3pt}
	\resizebox{\linewidth}{!}{   
		\begin{tabular}{c|cccccc}
			\toprule
			\textbf{Prompt} & AG $\uparrow$ & CE $\downarrow$ & EI $\uparrow$ & EN $\uparrow$ & Qcb $\uparrow$ & SF $\uparrow$ \\
			\midrule
			No Prompt  & 4.183  & 1.016  & 47.166  & 6.215  & 0.401  & 11.694 \\
			Simplified Prompt  & 4.431  & 0.610  & 48.687  & 6.613  & 0.419  & 12.186 \\
			Our Prompt  & \best{4.518}  & \best{0.596}  & \best{48.784}  & \best{6.639}  & \best{0.435}  & \best{12.554} \\
			\bottomrule
		\end{tabular}
	}
\end{table}

\section{Evalution Metrics}
We use six widely recognized image fusion evaluation metrics to objectively assess the quality of the fusion results. Among them, the average gradient (AG) \cite{55} measures the edge intensity of the image, indicating that the fused image can effectively restore the boundary of the target object. The contrast entropy (CE) \cite{58} measures the distribution of contrast in an image. The edge intensity (EI) \cite{59} measures the clarity and strength of edges in the image. The image entropy (EN) \cite{56} measures the complexity of details in the image. A higher image entropy indicates that the image contains more information and has more variations. The quality consistency based on fusion (Qcb) \cite{60} measures the quality consistency between the fused image and the reference image. The spatial frequency (SF) \cite{57} measures the spatial frequency of the image to ensure a better visual effect and clearer details in the fused image.

\section{Complexity Analysis}
In this section, we conduct a complexity analysis of the proposed method and compare it with several infrared-visible fusion methods that rely on infrared image generation. Table \ref{tab4} presents the complexity comparison results and inference time for each image among different methods. Compared with other pixel-level generation fusion methods, our method shows a significantly lower number of learnable parameters, moderate FLOPs and inference time, while maintaining superior performance. This advantage stems from our feature-level infrared modal inference and fusion framework, which eliminates the need for generating infrared images, thereby reducing computational complexity.

\begin{table}[htbp]
	\centering
	\caption{Complexity analysis of fusion models with missing infrared modality. All parameters listed in the table represent the learnable parameters of the models. The FLOPs and inference time for each model are calculated using input images of size $256\times256$. 'E' denotes the EGGAN method, while 'P' denotes the PID method.}
	\label{tab4}
	\tiny               
	\setlength{\tabcolsep}{2.8pt}   
	\renewcommand{\arraystretch}{0.95}  
	
	\resizebox{\columnwidth}{!}{%
		\begin{tabular}{l|c|c|c|c}
			\toprule
			\textbf{Model} & \textbf{Param (M)} & \textbf{FLOPs (G)} & \textbf{Time (s)} & \textbf{Qcb} $\uparrow$ \\
			\midrule
			E+U2Fusion   & 48.064 & 710.385 & 0.103 & 0.368 \\
			E+TarDAL     & 46.892 & 96.768  & 0.052 & 0.304 \\
			E+CDDFuse    & 47.784 & 194.176 & 0.059 & 0.429 \\
			E+LRRNet     & 46.644 & 80.347  & 0.026 & 0.390 \\
			E+IVFSWR     & 60.098 & 936.753 & 0.176 & 0.422 \\
			E+CoCoNet    & 55.709 & 118.866 & 0.075 & 0.409 \\
			E+EMMA       & 48.111 & 85.935  & 0.038 & 0.427 \\
			E+TIMFusion  & 47.827 & 111.154 & 0.053 & 0.396 \\
			E+DCEvo      & 48.601 & 272.027 & 0.059 & 0.432 \\
			E+SAGE       & 46.731 & 212.942 & 0.062 & 0.403 \\
			\midrule
			P+U2Fusion   & 242.895 & 2,921.78 & 11.066 & 0.388 \\
			P+TarDAL     & 241.723 & 2,308.16 & 11.015 & 0.282 \\
			P+CDDFuse    & 242.614 & 2,405.57 & 11.022 & 0.403 \\
			P+LRRNet     & 241.475 & 2,291.74 & 10.989 & 0.408 \\
			P+IVFSWR     & 254.929 & 3,148.14 & 11.139 & 0.377 \\
			P+CoCoNet    & 250.541 & 2,330.26 & 11.038 & 0.433 \\
			P+EMMA       & 242.776 & 2,297.33 & 11.002 & 0.390 \\
			P+TIMFusion  & 242.658 & 2,322.54 & 11.026 & 0.404 \\
			P+DCEvo      & 243.431 & 2,483.42 & 11.032 & 0.412 \\
			P+SAGE       & 241.562 & 2,424.33 & 11.025 & 0.339 \\
			\midrule
			\textbf{Ours} & \textbf{21.798} & \textbf{542.001} & \textbf{5.793} & \textbf{0.435} \\
			\bottomrule
	\end{tabular}} \vspace{-3mm}
\end{table}

\section{More experimental results}
To further evaluate the performance of our method, we conduct additional comparative experiments under missing-IR modility. Specifically, we first employ pixel-level infrared generation methods to reconstruct the missing infrared images and then fused the generated infrared images with visible images. The selected infrared generation methods include PID \cite{49}, a GAN-based approach EGGAN \cite{50}, and , a diffusion-based approach. The fusion process still adopts the ten mainstream fusion methods used in the main paper.  Our method does not generate infrared images. lnstead, it directly infers infrared features from the input visible images and produces enhanced images through the adaptive fusion of infrared andvisible features.

\begin{figure*}[t!]
	\centering
	\includegraphics[width=0.9\textwidth, height=0.95\textheight]{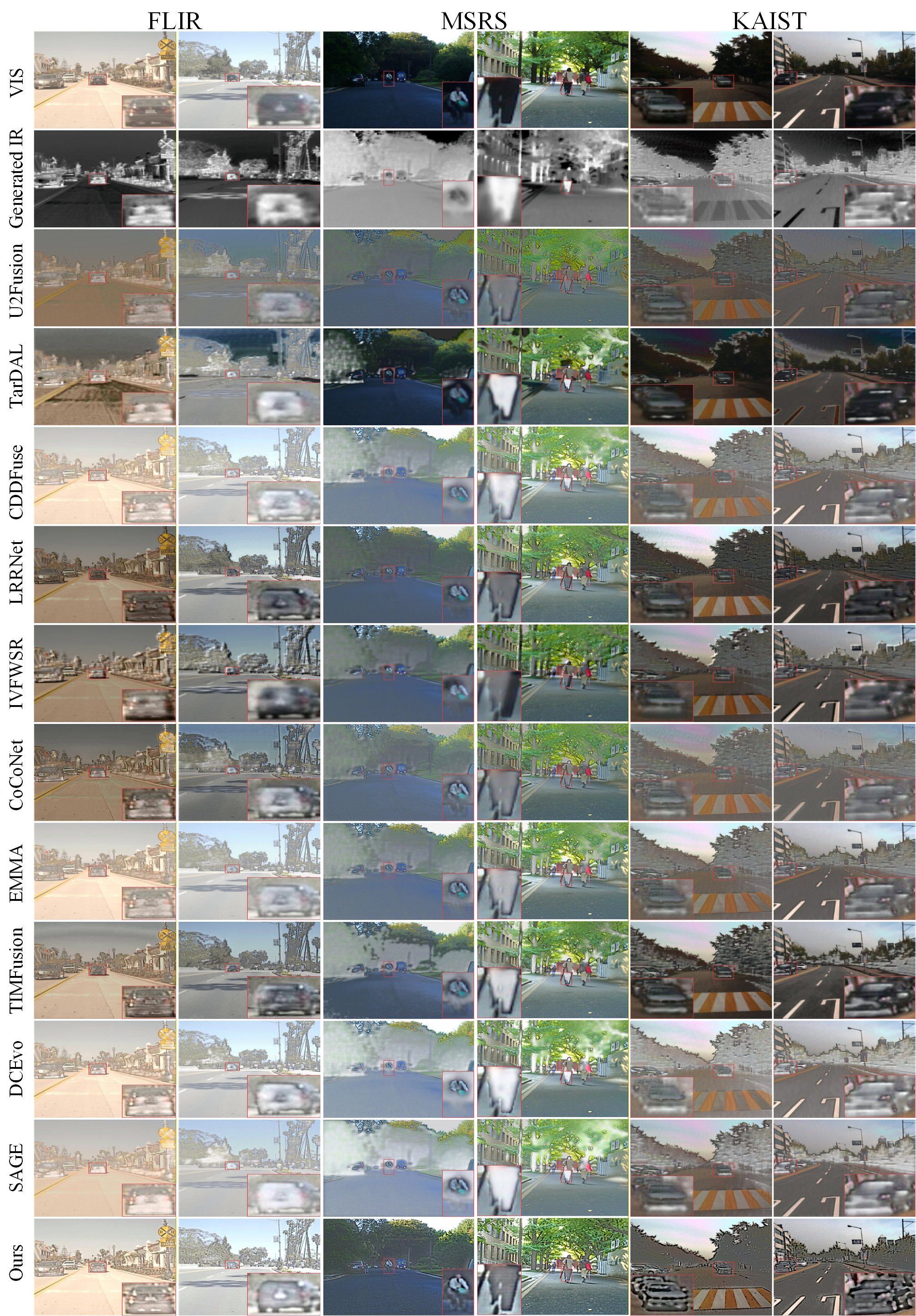}
	\caption{Visual comparison of different fusion methods on the FLIR, MSRS, and KAIST datasets, where the infrared images are generated from visible images using the EGGAN method.}\vspace{-5mm}
	\label{f2}
\end{figure*}

\begin{figure*}[t!]
	\centering
	\includegraphics[width=0.9\textwidth, height=0.95\textheight]{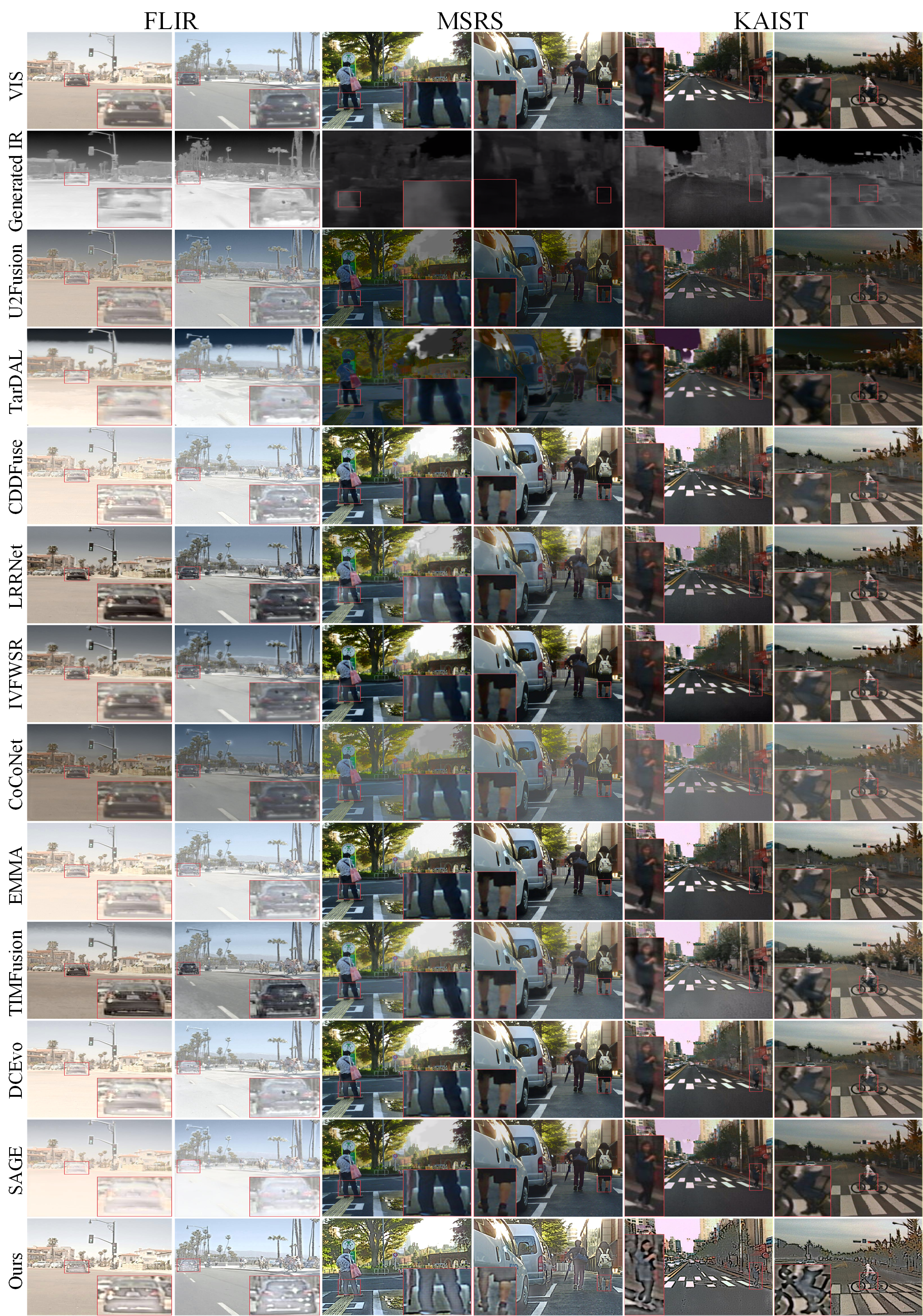}
	\caption{Visual comparison of different fusion methods on the FLIR, MSRS, and KAIST datasets, where the infrared images are generated from visible images using the PID method.}\vspace{-5mm}
	\label{f3}
\end{figure*}

As illustrated in Figures \ref{f2} and \ref{f3}, the qualitative results show that our method achieves an optimal balance among detail fidelity, thermal-target enhancement, and overall visual naturalness. The proposed approach effectively preserves the thermal information while fully integrating the structural and textural features of the visible image. In contrast, methods based on infrared generation suffer from modality imbalance, and their fusion results exhibit issues such as blurring, ghosting, and excessively bright or dark regions. The quantitative results in Tables \ref{tab5} and \ref{tab6} further validate that our method delivers superior performance across all evaluation metrics. These resluts clearly confirm that, for infrared–visible image fusion under missing infrared modality, our approach achieves state-of-the-art performance.

\section{Limitations and Future Work}
Although the proposed method is effective, it still has certain limitations. The method primarily addresses the catastrophic effects of missing infrared images but cannot fully replace the role of infrared images in challenging imaging environments. For example, under extreme weather conditions such as heavy fog, visible light imaging devices often fail to capture key targets, making it difficult to reliably infer significant infrared information from the visible images. This limitation constrains the upper bound of the fusion framework. In future work, we will further investigate fusion paradigms under missing infrared conditions, with a focus on enhancing the visual representation of visible images in uncontrollable imaging environments and reducing the uncertainty introduced by variations in imaging conditions and infrared feature inference. Additionally, we plan to incorporate more robust priors and learnable constraints to improve the method's generalization and stability in complex scenarios.

\begin{table*}[t]
	\centering
	\scriptsize
	\caption{Quantitative comparison of different fusion methods on the FLIR, MSRS, and KAIST datasets, where the infrared images are generated from visible images using the EGGAN method. The best and second-best performances are highlighted with \best{Red} and \second{Blue} backgrounds, respectively.} \vspace{-2mm}
	\renewcommand{\arraystretch}{1.2}
	\setlength{\tabcolsep}{2pt}  
	\resizebox{0.95\textwidth}{!}{
		\begin{tabular}{c|cccccc|cccccc|cccccc}
			\toprule
			\textbf{Datasets} & \multicolumn{6}{c|}{\textbf{MSRS}} & \multicolumn{6}{c|}{\textbf{FLIR}} & \multicolumn{6}{c}{\textbf{KAIST}}\\
			\midrule
			\textbf{Methods} & AG $\uparrow$ & CE $\downarrow$ & EI $\uparrow$ & EN $\uparrow$ & Qcb $\uparrow$ & SF $\uparrow$ & AG $\uparrow$ & CE $\downarrow$ & EI $\uparrow$ & EN $\uparrow$ & Qcb $\uparrow$ & SF $\uparrow$ & AG $\uparrow$ & CE $\downarrow$ & EI $\uparrow$ & EN $\uparrow$ & Qcb $\uparrow$ & SF $\uparrow$ \\
			\midrule
			U2Fusion  & 3.289  & 3.309  & 34.980  & 5.447  & 0.327  & 9.310 & 2.320  & 1.453  & 24.107  & 5.045  & 0.368  & 6.517  & 2.425  & 2.734  & 25.699  & 5.489  & 0.388  & 6.259 \\
			TarDAL   & 3.205  & 3.458  & 34.550  & 6.640  & 0.363  & 8.602 & 2.341  & 2.756  & 25.123  & 6.536  & 0.304  & 6.898  & 1.916  & \best{1.728}  & 20.399  & 6.197  & 0.382  & 5.807 \\
			CDDFuse  & 3.644  & \second{2.839}  & 38.606  & 6.513  & 0.318  & 11.581 & 3.004  & \best{0.515}  & 31.810  & 6.406  & 0.429  & 8.746  & 2.944  & 4.315  & 31.419  & 6.225  & 0.374  & 8.835 \\
			LRRNet  & 3.979  & 4.924  & 42.225  & 6.617  & 0.340  & 12.258  & 3.905  & 1.993  & 41.323  & \best{6.866}  & 0.390  & \second{11.946}  & 3.086  & 2.443  & 33.250  & 6.642  & 0.418  & 8.862 \\
			IVFWSR   & 3.229  & 4.218  & 34.902  & 6.461  & 0.378  & 8.608 & 3.000  & 3.704  & 32.530  & 6.349  & 0.422  & 7.818  & 2.635  & 2.631  & 28.661  & 6.462  & 0.414  & 6.959 \\
			CoCoNet  & 3.930  & \best{2.664}  & 41.488  & 6.312  & \second{0.385}  & 11.626 & \second{3.948}  & 3.267  & \second{41.592}  & 6.363  & 0.409  & 11.136  & 2.716  & 3.055  & 28.735  & 6.006  & 0.390  & 7.524 \\
			EMMA  & \second{4.466}  & 3.236  & \second{47.469}  & 6.608  & 0.366  & 13.207 & 3.028  & 0.691  & 32.297  & 6.467  & 0.427  & 8.689  & 3.178  & 2.969  & 34.234  & 6.405  & 0.414  & 8.941 \\
			TIMFusion  & 3.887  & 5.072  & 41.836  & \second{6.839}  & 0.356  & 11.165 & 3.892  & 2.938  & 41.172  & 6.571  & 0.396  & 11.900  & \second{3.479}  & 2.473  & \best{37.688}  & \second{6.894}  & \best{0.446}  & 9.279 \\
			DCEvo   & 4.170  & 4.718  & 44.500  & 6.814  & 0.308  & \second{13.240}  & 3.246  & \second{0.591}  & 34.602  & 6.532  & \second{0.432}  & 9.768  & 3.451  & 4.359  & 37.032  & 6.628  & 0.375  & \second{9.887} \\
			SAGE  & 4.150  & 3.057  & 44.114  & 6.777  & 0.327  & 12.560 & 2.724  & 1.160  & 28.303  & 5.966  & 0.403  & 8.138  & 3.376  & 2.986  & 35.924  & 6.608  & \second{0.430}  & 9.534 \\
			Ours    & \best{5.037} & 3.142 & \best{53.168} & \best{7.188} & \best{0.454} & \best{14.898} & \best{4.518} & 0.596 & \best{48.784} & \second{6.639} & \best{0.435} & \second{12.554} & \best{4.414} & \second{2.227} & \second{37.115} & \best{7.073} & 0.405 & \best{11.031} \\
			\bottomrule  
		\end{tabular}    \vspace{-1000mm}
		\label{tab5}
	}
\end{table*}

\begin{table*}[htbp]
	\centering
	\scriptsize
	\caption{Quantitative comparison of different fusion methods on the FLIR, MSRS, and KAIST datasets, where the infrared images are generated from visible images using the PID method. The best and second-best performances are highlighted with \best{Red} and \second{Blue} backgrounds, respectively.} \vspace{-2mm}
	\renewcommand{\arraystretch}{1.2}
	\setlength{\tabcolsep}{2pt}  
	\resizebox{0.95\textwidth}{!}{
		\begin{tabular}{c|cccccc|cccccc|cccccc}
			\toprule
			\textbf{Datasets} & \multicolumn{6}{c|}{\textbf{MSRS}} & \multicolumn{6}{c|}{\textbf{FLIR}} & \multicolumn{6}{c}{\textbf{KAIST}}\\
			\midrule
			\textbf{Methods} & AG $\uparrow$ & CE $\downarrow$ & EI $\uparrow$ & EN $\uparrow$ & Qcb $\uparrow$ & SF $\uparrow$ & AG $\uparrow$ & CE $\downarrow$ & EI $\uparrow$ & EN $\uparrow$ & Qcb $\uparrow$ & SF $\uparrow$ & AG $\uparrow$ & CE $\downarrow$ & EI $\uparrow$ & EN $\uparrow$ & Qcb $\uparrow$ & SF $\uparrow$ \\
			\midrule
			U2Fusion & 3.074  & 1.668  & 32.810  & 6.047  & 0.428  & 9.168 & 3.277  & 1.353  & 33.900  & 6.576  & 0.388  & 8.329 & 2.533  & \second{0.828}  & 27.040  & 6.180  & 0.459  & 6.911 \\
			TarDAL  & 1.938  & 2.342  & 20.524  & 6.030  & 0.378  & 5.654 & 2.516  & \second{0.643}  & 26.795  & \second{7.055}  & 0.282  & 6.860  & 1.937  & 1.268  & 20.667  & 6.302  & 0.427  & 5.766 \\
			CDDFuse  & 4.821  & 1.651  & 51.500  & \second{7.191}  & 0.466  & 14.142 & 3.837  & 0.726  & 39.804  & 6.517  & 0.403  & 10.254  & 3.349  & 1.361  & 35.909  & \second{7.064}  & 0.488  & 9.859 \\
			LRRNet  & 3.815  & 2.837  & 40.509  & 7.049  & 0.445  & 11.150 & \second{3.970}  & 1.798  & 42.038  & 6.955  & 0.408  & 11.154  & 3.125  & 1.432  & 33.685  & 7.017  & 0.439  & 8.740 \\
			IVFWSR  & 3.994  & \second{1.342}  & 43.035  & 7.046  & \second{0.470}  & 11.054 & 2.976  & 1.266  & 31.999  & \best{7.180}  & 0.377  & 7.546  & 2.878  & 0.857  & 31.010  & 6.962  & 0.442  & 8.313 \\
			CoCoNet  & 3.357  & \best{0.801}  & 35.271  & 6.477  & 0.418  & 10.386 & 2.456  & 1.848  & 25.333  & 6.417  & \second{0.433}  & 6.589  & 2.200  & 1.073  & 23.195  & 6.240  & 0.456  & 6.444 \\
			EMMA   & \second{4.904}  & 1.353  & \second{52.540}  & 7.187  & 0.440  & \second{14.253} & 3.150  & 0.745  & 33.517  & 6.280  & 0.390  & 8.830  & 3.334  & \best{0.813}  & 36.033  & 6.979  & \second{0.495}  & 9.655 \\
			TIMFusion  & 3.952  & 3.928  & 42.063  & 6.936  & 0.374  & 11.708 & 3.967  & 1.541  & \second{42.121}  & 6.982  & 0.404  & \second{11.572}  & 3.430  & 1.877  & \second{36.806}  & 7.004  & 0.434  & 9.509 \\
			DCEvo    & 4.722  & 1.983  & 50.492  & \best{7.231}  & \best{0.476}  & 13.974 & 3.468  & 0.812  & 36.719  & 6.531  & 0.412  & 9.516  & \second{3.439}  & 1.209  & 36.676  & 6.998  & \best{0.497}  & \second{10.381} \\
			SAGE     & 4.447  & 1.867  & 47.126  & 7.005  & 0.436  & 13.246 & 2.690  & 0.697  & 27.959  & 6.379  & 0.339  & 7.818  & 3.234  & 1.035  & 34.370  & 6.805  & 0.416  & 9.574 \\
			Ours    & \best{5.037} & 3.142 & \best{53.168} & 7.188 & 0.454 & \best{14.898} & \best{4.518} & \best{0.596} & \best{48.784} & 6.639 & \best{0.435} & \best{12.554} & \best{4.414} & 2.227 & \best{37.115} & \best{7.073} & 0.405 & \best{11.031} \\
			\bottomrule  
		\end{tabular}  
		\label{tab6}
	}
\end{table*}

\end{document}